\pdfoutput=1

\documentclass[11pt]{article}

\usepackage[final]{acl}

\usepackage{times}
\usepackage{latexsym}
\usepackage{amsmath}

\usepackage[T1]{fontenc}

\usepackage[utf8]{inputenc}

\usepackage{inconsolata}

\usepackage{microtype}
\usepackage{multirow}
\usepackage{url}            
\usepackage{booktabs}       
\usepackage{array, graphicx, tabularx}
\usepackage{xcolor} 
\usepackage[table]{xcolor}
\usepackage{tcolorbox}
\usepackage{fontawesome}
\usepackage{float}
\usepackage{wrapfig}
\usepackage{caption}

\newcommand{\ourmethod}{\textit{FM$^2$DS}}
\newcommand{\benchmark}{\textit{M$^2$QA-Bench}}

\title{
    \raisebox{-1.1ex}{\includegraphics[width=1cm]{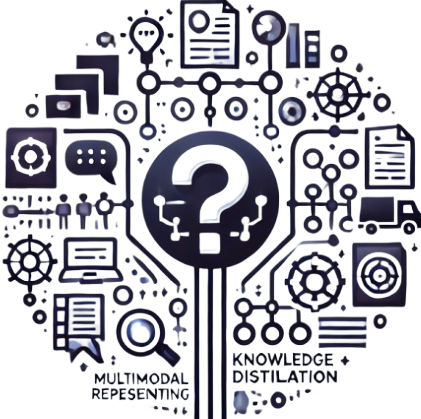}}\,%
    \ourmethod: Few-Shot Multimodal Multihop Data Synthesis \\ with Knowledge Distillation for Question Answering
}

\author{
    Amirhossein Abaskohi$^{1,2}$, 
    Spandana Gella$^{2}$, 
    Giuseppe Carenini$^{1}$, 
    Issam H. Laradji$^{1,2}$ \\
    \\
    $^1$ Department of Computer Science The University of British Columbia \\
    V6T 1Z4, Vancouver, BC, Canada \\ \\
    $^2$ ServiceNow Research 
}

\begin{document}

\newfloat{prompt}{htbp}{lop} 
\floatname{prompt}{Prompt} 

\definecolor{prompt}{RGB}{223, 223, 192}
\definecolor{prompt-frame}{RGB}{137, 137, 90}
\definecolor{prompt-title}{RGB}{117, 117, 77} 

\newtcolorbox{promptbox}[2][]{colback=prompt, colframe=prompt-frame,
  fonttitle=\bfseries,
  title=\faDatabase \space #2,
  top=2mm,
  left=2mm,
  coltitle=black,
  colbacktitle=prompt-title,
  colframe=prompt-frame,
  rounded corners, 
  #1
}

\definecolor{example}{RGB}{226, 237, 252}
\definecolor{example-frame}{RGB}{80, 120, 160}

\newfloat{example}{htbp}{lop} 
\floatname{example}{Example} 

\newtcolorbox{examplebox}[1][]{colback=example, colframe=example-frame,
  fonttitle=\bfseries,
  top=2mm,
  left=2mm,
  coltitle=black,
  colframe=example-frame,
  rounded corners, 
  #1
}

\newcounter{examplewrap}
\renewcommand{\theexample}{\arabic{examplewrap}}

\newcommand{\wrappedexample}[4][]{%
    \refstepcounter{examplewrap}%
    \begin{wrapfigure}{#2}{0.45\textwidth}
        \vspace{-\intextsep}
        \begin{minipage}{0.45\textwidth}
            \begin{examplebox}[#1]
                #3
            \end{examplebox}
            \captionof{example}{#4}
            \label{example:\theexample}
        \end{minipage}
    \end{wrapfigure}
}

\maketitle

\definecolor{softblue}{RGB}{220,240,255}
\definecolor{softorange}{RGB}{255,240,220}
\definecolor{softgreen}{RGB}{220,255,220}

\begin{abstract}
Multimodal multihop question answering  (MMQA) requires reasoning over images and text from multiple sources, an essential task for many real-world applications. Despite advances in visual question answering, the multihop setting remains underexplored due to the lack of quality datasets. Existing methods focus on single-hop, single-modality, or short texts, limiting real-world applications like interpreting educational documents with long, multimodal content. To fill this gap, we introduce \ourmethod, the first framework for creating a high-quality dataset for MMQA. Our approach consists of a five-stage pipeline that involves acquiring relevant multimodal documents from Wikipedia, synthetically generating high-level questions and answers, and validating them through rigorous criteria to ensure data quality. We evaluate our methodology by training models on our synthesized dataset and testing on two benchmarks: MultimodalQA and WebQA. Our results demonstrate that, with an equal sample size, models trained on our synthesized data outperform those trained on human-collected data by 1.9 in exact match (EM) score on average.  Additionally, we introduce \benchmark\ with 1k samples, the first benchmark for MMQA on long documents, generated using \ourmethod\  and refined by human annotators\footnote{Code is publicly available at: \url{https://github.com/ServiceNow/FM2DS}.}. 
\end{abstract}

\section{Introduction}

\begin{figure}
    \centering
    \includegraphics[width=\linewidth]{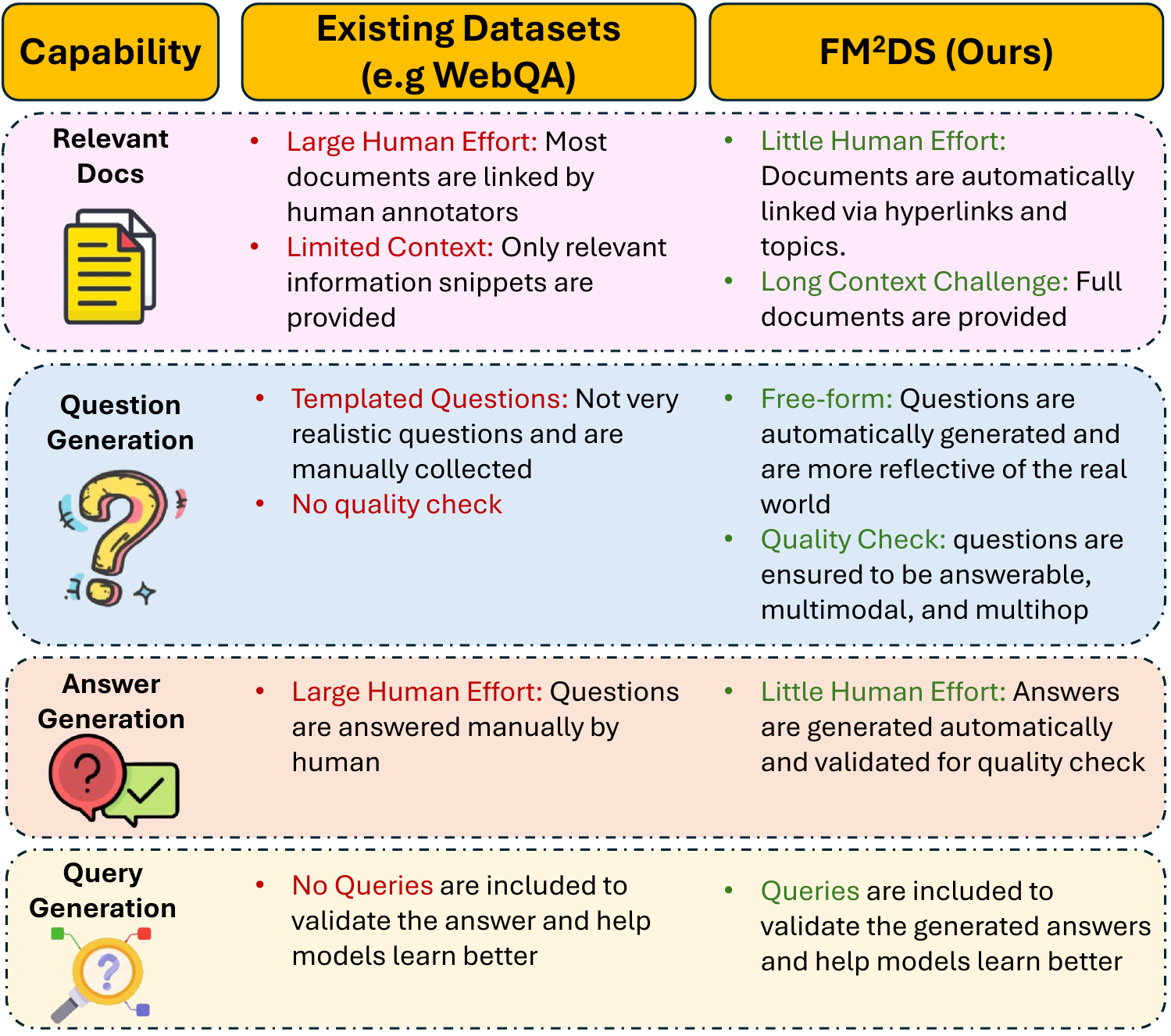}
    \caption{Unlike traditional datasets that rely on human annotators, templates, or snippets, \ourmethod\ is fully automated, using long documents as sources and applying validation to ensure questions are answerable, multimodal, and multihop.}
    \label{fig:teaser}
\end{figure}

Multimodal multihop question answering (MMQA) involves answering complex questions by integrating information from text, images, and tables. 
In real-world applications such as interpreting medical documents, this challenge is amplified by the need to reason over long, multimodal content. 
Current methodologies in MMQA typically leverage in-context learning methods, prompting large vision language models (LVLMs) to retrieve relevant information from multimodal sources \citep{tejaswi2024rare} and then perform reasoning \citep{10.1145/3581783.3611964}. However, these models often demand significant computational resources due to their large parameter counts, making them costly to deploy even during inference \citep{ye2024voco}. This limitation emphasizes the need for more efficient frameworks that can operate effectively with minimal annotated data. A practical solution is to use a smaller model capable of both retrieving the necessary information from sources and performing reasoning. This can be achieved by fine-tuning the model on a MMQA dataset. Fine-tuning enables domain specialization, allowing the model to adapt to specific areas of interest. It also requires significantly less compute and memory compared to large commercial models. Finally, this approach reduces privacy concerns by enabling training on sensitive domains such as legal, medical, or proprietary data that models like GPT-4o \cite{openai2024gpt4technicalreport} cannot access. Existing datasets often rely on short snippets or repetitive templates, limiting generalizability to complex settings with long texts and multiple modalities \citep{WebQA21, talmor2021multimodalqa, jiang2024understanding, chen2024we}. Additionally, creating new similar datasets is challenging, requiring extensive human annotation \citep{lu2022learn,chen2023can}. 

In this work, we propose \ourmethod, a novel data synthesis framework designed specifically for MMQA over \textbf{long documents}. 
Our approach synthesizes MMQA data from documents that are interconnected through various relationships, such as thematic similarities or sequential events. This framework leverages naturally occurring document relationships and requires minimal hand-crafted data, thereby broadening the range of reasoning types used in question generation.

As illustrated in Figure \ref{fig:teaser}, \ourmethod\ enables the generation of non-templated question-answer pairs based on full documents rather than brief information snippets. 
The data generated by our method
\ourmethod\ includes \textbf{query} component - \textbf{a step-by-step guide for retrieving relevant information from multiple documents} - enabling smaller models trained on this synthesized data to learn how to tackle complex questions in a manner similar to larger models.
This methodology allows users to create a custom MMQA dataset with fewer than ten human-annotated samples, thereby facilitating the fine-tuning of smaller LVLMs for specific applications.

\ourmethod\ leverages Wikipedia’s extensive knowledge base and hyperlink structure to select document pairs with shared topical relevance or hyperlink connections and prompt LVLMs to perform question generation, question answering, and query generation. 
We incorporate validation steps to enhance the quality of the generated data and discard any outputs that are factually incorrect. 
Through empirical evaluation on established MMQA benchmarks, we show that \ourmethod\ significantly improves model performance, achieving on average a 1.9 exact match (EM) score improvement across two benchmarks: MultimodalQA and WebQA. 

Our key \textbf{contributions} are: \textbf{(I)} introducing a new framework for synthesizing high-quality MMQA training data for LVLMs; \textbf{(II)} using a robust validation pipeline to ensure data quality; \textbf{(III)} introducing a challenging MMQA benchmark requiring reasoning over multiple modalities and sources; and \textbf{(IV)} showing that models fine-tuned on our synthetic data outperform those trained on human-labeled datasets, advancing MMQA while reducing manual effort.

\section{Related Work}
\label{sec:related-work}

Within the Question Answering (QA) literature, synthesis of training data has been predominantly focused on unimodal (text-only) scenarios. We review various similar works that have established the foundation for our work in few-shot data synthesis.

\paragraph{Unimodal Data Synthesis}
Synthetic data is increasingly used for model training. \citet{he-etal-2022-generate} show that combining labeled and synthetic text from language models (LMs) improves NLP performance. Entire synthetic datasets have also been created for tasks like classification \citep{Tsui_AnyClassifier_2024}, with \citet{li2023syntheticdatagenerationlarge} demonstrating GPT-3.5’s effectiveness in generating reliable classification data. Similarly, \citet{chen-etal-2024-shot} show that synthetic data can significantly boost small models on multi-hop QA with minimal human annotation.

\paragraph{Multimodal Data Synthesis} 

\begin{figure*}[!ht]
    \centering
    \includegraphics[width=0.97\textwidth]{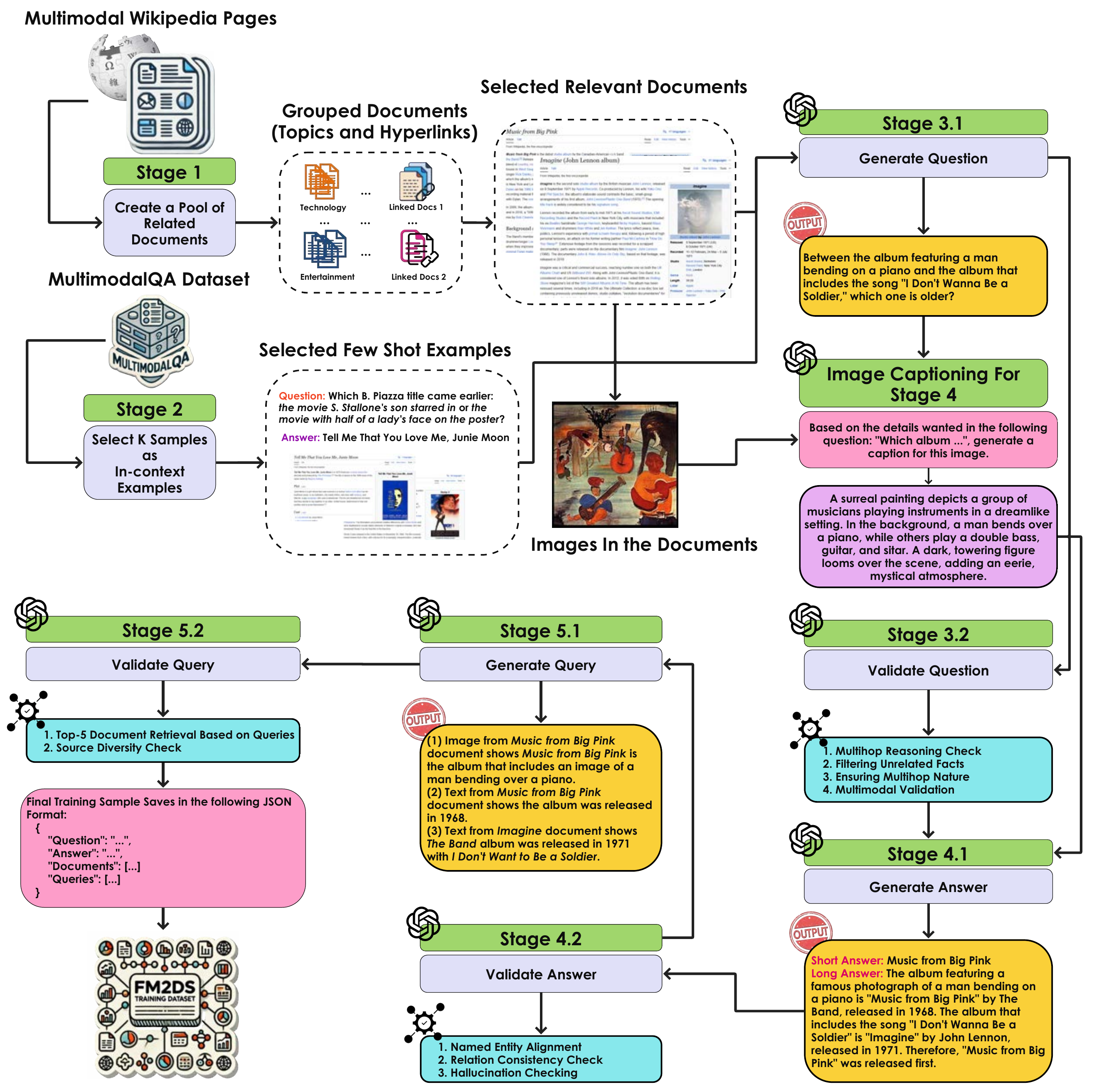}
    \caption{The \ourmethod\ pipeline consists of five stages for generating high-quality multihop multimodal QA samples. In Stage 1, a pool of related Wikipedia documents is retrieved by leveraging topic similarity and hyperlink connections to ensure contextual richness. Stage 2 selects few-shot in-context examples from the MultiModalQA dataset \cite{talmor2021multimodalqa} to guide generation. Stage 3 focuses on question generation (3.1) and validation (3.2), ensuring questions require multihop reasoning, are answerable, and grounded in both text and images. Stage 4 generates (4.1) and validates (4.2) answers through named entity alignment, relation consistency, and hallucination checks. Finally, Stage 5 generates (5.1) and validates (5.2) retrieval queries to collect diverse and relevant supporting documents. The resulting samples are saved in a structured format for use in MMQA training and evaluation.}
    \label{fig:pipeline}
\end{figure*}

Research on multimodal data synthesis with LVLMs remains limited, with most efforts focused on generating new data from model's pre-trained knowledge. \citet{zhang2024multimodal} synthesize abstract images with reasoning tasks, while \citet{mehta2024fake} generate multimodal data using unimodal models for pre-training. In MMQA, \citet{wu-etal-2024-synthetic} propose SMMQG, which uses multimodal RAG to generate questions from short snippets, focusing on multimodality rather than multihop reasoning. In contrast, \ourmethod\ uses full multimodal documents, resulting in a more challenging dataset with diverse multihop questions that better reflect real-world tasks. Moreover, while SMMQG is confined to predefined question types, \ourmethod\ enables large-scale generation and supports knowledge distillation for smaller models through step-by-step queries that guide complex multi-document reasoning.

\section{Proposed Method: \ourmethod}
\label{sec:method}
Our five-stage pipeline for \ourmethod\ (Figure~\ref{fig:pipeline}) synthesizes high-quality multimodal QA pairs. It begins by grouping documents via topic matching and Wikipedia hyperlinks, followed by few-shot sample selection, question synthesis, answer generation, and query construction, each with their built-in validation. See Appendix \ref{appendix:samples} for examples.

\subsection{Stage 1: Creating a Pool of Related Documents}
We collect relevant documents from Wikipedia using the WikiWeb2M dataset \citep{burns2023wiki}, which includes nearly 2 million pages. Documents are linked via two methods: hyperlinks and latent topics identified through multimodal topic modeling with the Multimodal-Contrast model \citep{gonzalez2024neural}. Since Multimodal-Contrast can not handle long documents, we split each document into shorter segments containing at most one image, apply topic modeling to each segment, then merge the results and remove duplicates. This combination captures both clear and subtle relationships across documents, integrating textual and visual information.

\begin{table*}[!t]
    \centering
    \small
    \begin{tabular}{>{\columncolor{softblue}}m{0.6\textwidth} >{\columncolor{softblue}}m{0.3\textwidth}}
    \textbf{Question} & \textbf{Type} \\
    \hline
    In what year did Mike Tyson become the youngest heavyweight champion, and who is the president of the United States? & Unrelated Facts \\
    \end{tabular}
    
    \vspace{2pt}
    
    \begin{tabular}{>{\columncolor{softorange}}m{0.6\textwidth} >{\columncolor{softorange}}m{0.3\textwidth}}
    \textbf{Question} & \textbf{Type} \\
    \hline
    In what year did Mike Tyson become the youngest heavyweight champion, and who was the president of the United States at that time? & Related Facts, Open-ended \\
    \end{tabular}
    
    \vspace{2pt}
    
    \begin{tabular}{>{\columncolor{softgreen}}m{0.6\textwidth} >{\columncolor{softgreen}}m{0.3\textwidth}}
    \textbf{Question} & \textbf{Type} \\
    \hline
    Who was the president of the United States when Mike Tyson became the youngest heavyweight champion? & Concise Multihop Question \\
    \end{tabular}
    
    \caption{Examples of factual questions with varying degrees of relevance and conciseness, demonstrating progression from unrelated to concise multihop reasoning.}
    \label{tab:example-questions}
\end{table*}

\subsection{Stage 2: Creating Few-Shot Samples}
We sample multihop questions from the MultiModalQA dataset \citep{talmor2021multimodalqa}, which requires reasoning across text, images, and tables. As our samples are based on full documents rather than short information snippets like in MultimodalQA, we crawled the complete Wikipedia HTML pages using the entity links provided in MultimodalQA, which are associated with the dataset's examples. We then compiled few-shot samples using these full HTML pages, complete with images and tables, paired with their corresponding questions. We randomly select up to three samples for question generation in our experiments.

\subsection{Stage 3: Question Generation and Validation}
\label{subsec:question_generation}

\paragraph{Question Generation}

We use GPT-4-turbo \citep{openai2024gpt4technicalreport} to generate multihop, multimodal questions from few-shot samples based on MultiModalQA few-shot examples. Due to context limitations, inputs are limited to grouped sets containing 2 or 3 documents. Our prompt (see Appendix \ref{appendix:prompts}) is designed to ensure that questions require reasoning across all documents and at least two modalities, avoiding unrelated combinations such as: “How did Einstein contribute to relativity and when was Princeton established?"; a question that spans multiple documents, but lacks meaningful multihop reasoning.


\paragraph{Question Validation} Our framework includes validation stages to ensure questions meet multihop and multimodal criteria. 
While the model was prompted to avoid simple concatenations, we further evaluated this aspect.

We used LLama-3.1-8B \citep{dubey2024llama3herdmodels} to decompose questions and check if parts could be answered with a single document. If all parts of the question were answerable with a single document, we discarded such question that include unrelated facts (see \colorbox{softblue}{\textit{Unrelated Facts}} example in Table \ref{tab:example-questions}). Otherwise, we retained only the the parts of the questions that required information from multiple documents to ensure the revised question met the multihop criteria. However, a potential issue was that, even when the facts were related, the questions could still become open-ended, requiring explanations or combined answers (see example \colorbox{softorange}{\textit{Related Facts, Open-ended}} in Table \ref{tab:example-questions}). In order to follow the standard of question answering, and make the evaluation process easier, we used GPT-4o to rephrase the question without conjunctions while maintaining its multihop nature, resulting in \colorbox{softgreen}{\textit{Concise Multihop Question}} (Table \ref{tab:example-questions}).

Another key step in validation was ensuring the questions were truly multimodal. After verifying that a question was multihop, we tested whether it remained answerable when the documents were limited to a single modality (e.g., text-only, image-only, or table-only). Using GPT-4o (refer to Section \ref{subsec:stage4} for details), we checked if the question could be answered with just one modality. If so, we discarded it, as it failed to meet the multimodal requirement. This step helped refine the dataset to include only questions that genuinely required reasoning across multiple modalities and documents.

\subsection{Stage 4: Answer Generation and Validation}
\label{subsec:stage4}

\paragraph{Answer Generation}
We used GPT-4o to generate concise answers from multiple documents, including text and images. The model was instructed to provide a long answer and a short answer with only key information and no extra explanation. To help the model focus on specific details of images in the given documents to answer the multimodal question, we include question-related captions for the images. For example, if the question asks about the geometric shapes in an image (see Figure \ref{fig:q2}), the model generates a caption describing the shapes. This makes it easier for the model to answer the question accurately.

\paragraph{Answer Validation}
We validated answers using named entity recognition (NER) and relation extraction, following prior work \citep{rajpurkar-etal-2018-know, fabbri-etal-2022-qafacteval}. NER ensured key entities and numbers in the answer matched the documents, while relation extraction verified that entity relationships were consistent with the source content (via Spacy \citep{wu2019enrichingpretrainedlanguagemodel}). For including image content, we used the same question-related caption generated by GPT-4o (e.g., noting a building’s color if relevant to the question) similar to answer generation. To reduce hallucinations, we prompted GPT-4o five times and accepted answers only if all outputs (5/5) agreed. To evaluate the effectiveness of our answer validation process, we conducted a human study to assess the quality of the filtered questions and answers. The results of this evaluation are presented in Section \ref{sec:human_eval}.

\subsection{Stage 5: Query Generation and Validation}

\paragraph{Query Generation}
We generate queries using GPT-4o based on the question-answer pairs and related documents to enhance retrieval effectiveness. These queries guide the smaller model trained on \ourmethod-generated data to retrieve specific and relevant information, improving its ability to answer questions accurately. By narrowing down the content, we can extract key details such as named entities, relationships, and contextual cues aligned with the question. This targeted approach ensures that the generated answers are not only concise and accurate, but also directly grounded in evidence from the documents.

\paragraph{Query Validation}

To validate the queries, we used MuRAG \citep{chen-etal-2022-murag}, which encodes text and images into a shared embedding space for multimodal retrieval. For each generated query, we retrieved the top-5 documents retrieved by MuRAG. If more than one of the original source documents used to generate the question appeared in the top-5, the query was considered well-formed. This process ensures the query effectively captures diverse, relevant information and can help teach smaller models how to retrieve supporting evidence for answering questions.

\section{Proposed Benchmark: \benchmark}

We introduce \benchmark, a benchmark of 1k diverse Q\&A pairs to evaluate LVLMs on complex MMQA with full documents. Unlike templated datasets \citep{talmor2021multimodalqa}, questions are varied and challenging (see Appendix \ref{appendix:benchmark_info} and Appendix \ref{appendix:zero_context_baseline} for details on diversity and complexity). Answering requires cross-modal reasoning and information extraction from full documents, including images and tables. See Table \ref{tab:benchmark_stats} for key statistics (more in Appendix \ref{appendix:benchmark_info}) and Appendix \ref{appendix:samples} for samples generated by \ourmethod\ for \benchmark. To create this benchmark, we used the \ourmethod\ pipeline to generate 1,200 samples, which were verified by three annotators for correctness, multihop reasoning, multimodality, and answer accuracy. Each sample was scored 1 (valid) or 0 (invalid). This annotation required minimal human effort (2.2 min/question on average) due to structured queries. Samples averaging below 0.75 were removed, leaving 1,142 (i.e removing only 5\% of the total); we then randomly selected 1,000 for the benchmark to ensure consistency in evaluation and reduce potential sampling bias. Annotator agreement (\textbf{Fleiss' Kappa} \citep{fleiss1971measuring}) was \textbf{0.83}.

\begin{table}[!h]
    \centering
    \resizebox{\linewidth}{!}{%
    \begin{tabular}{l|c}
    \hline
    \textbf{Statistic} & \textbf{Value} \\
    \hline
    Image Modality Percentage & 73.6\% \\
    Table Modality Percentage & 89.6\% \\
    Both Image and Table Modality Percentage & 63.6\% \\
    Average Question Length (Word) & 23.77  \\
    Average Answer Length (Word) & 1.95  \\
    Average Source Documents Per Question & 2.29 \\
    \hline
    \end{tabular}
    }
    \caption{Key statistics of the proposed multimodal multihop question answering benchmark.}
    \label{tab:benchmark_stats}
\end{table}

\section{Experiments and Results}
\label{sec:results}

This section compares our synthesized dataset to human-annotated ones. All experiments used \textit{one in-context example} during synthesis (see Appendix \ref{appendix:different_shots} for effects of varying number in-context examples). GPT-4o was used in the pipeline (Appendix \ref{appendix:other_generative_models} shows results with other LVLMs). Models were evaluated using Exact Match (EM) for accuracy and F1 for partial match quality. Further experimental details can be found in Appendix \ref{appendix:exp_settings}.

\subsection{Details of Synthesized Experimental Training Data}
\label{subsec:train_data_details}

We synthesize an 18k-sample training set with \ourmethod\ under the 3-shot setting, using a 20-example few-shot pool (10 from MultimodalQA, 10 from WebQA) to guide style and difficulty. Table~\ref{tab:training_stats} reports aggregate corpus statistics. The dataset is inherently multimodal: the majority of questions reference images, tables, or both. In most of our experiments, we use either a 5k or 10k subset of this training data to balance efficiency and performance. Larger subsets (\(>10\)k) are employed only in special cases where we aim to identify the minimum amount of synthesized data required to surpass training on the full ground-truth training set of the respective test benchmark. This threshold varies depending on the model and the evaluation dataset. To ensure training utility, we enforce basic validity checks during synthesis (e.g., modality availability, answerability, and cross-source consistency) and remove low-confidence or duplicate generations. These design choices yield a balanced, compact corpus that emphasizes multimodal reasoning without sacrificing clarity. See Appendix \ref{appendix:sample_cost} for data generation cost details.

\begin{table}[!h]
    \centering
    \resizebox{\linewidth}{!}{%
    \begin{tabular}{l|c}
    \hline
    \textbf{Statistic} & \textbf{Value} \\
    \hline
    Image Modality Percentage & 64.55\% \\
    Table Modality Percentage & 64.4\% \\
    Both Image and Table Modality Percentage & 42.2\% \\
    Average Question Length (Word) & 22.5  \\
    Average Answer Length (Word) & 1.94  \\
    Average Source Documents Per Question & 2.1 \\
    \hline
    \end{tabular}
    }
    \caption{Key statistics of the 18k synthesized training samples generated by \ourmethod\ under the 3-shot setting. While we generated 18k samples in total, most experiments use 5k or 10k subsets, with larger subsets employed only when testing the minimum required size to outperform training on the original ground-truth training data of the respective benchmarks.}
    \label{tab:training_stats}
\end{table}

\subsection{Training Details}
\label{subsec:training_details}

\paragraph{Structured Query Format for Knowledge Distillation.}  
To promote explicit and grounded multimodal reasoning, we train models in a structured query–answer format where each prediction requires the model to (I) identify the relevant modality (image, table, or both), (II) extract or cite supporting evidence, and (III) generate the final answer. Figures~\ref{fig:sample_1} and~\ref{fig:sample_2} in Appendix~\ref{appendix:samples} illustrate examples of queries. The model is supervised to generate both intermediate queries and final answers using a masked language modeling (MLM) loss, preventing shortcut learning and ensuring that reasoning is explicitly tied to content.

We align this framework with knowledge distillation from a stronger teacher model (GPT-4o). The teacher produces both structured queries (chain-of-thought style reasoning) and answers, which serve as distillation targets. A validation step filters out hallucinated or factually inconsistent teacher queries before use. The student model is then optimized to mimic the teacher’s reasoning and answer trajectories, similar in spirit to reasoning-focused distillation methods such as DeepSeek-R1-Distill~\cite{DBLP:journals/corr/abs-2501-12948}, but adapted for multimodal multihop QA.  

Queries are essential for guiding models to use the provided context, as without explicit multimodal grounding they struggle to answer reliably despite strong pretrained knowledge. As demonstrated in Appendix~\ref{appendix:zero_context_baseline}, removing supporting context causes sharp performance drops across all baselines, highlighting the importance of grounding multimodal multihop question answering.

\paragraph{Training Objective.}  
Given an input $(x, y, r)$ where $x$ is the multimodal context (including the question and documents), $r$ is the structured query (teacher reasoning), and $y$ is the ground-truth answer, the model is optimized with a joint objective:  
\[
\mathcal{L} = \mathcal{L}_{\text{CLM}}(r|x) + \mathcal{L}_{\text{CLM}}(y|x, r),
\]  
where $\mathcal{L}_{\text{CLM}}$ denotes the causal language modeling loss. The first term enforces generation of factually grounded queries, while the second supervises answer generation conditioned on both the input and queries. In the distillation setting, teacher outputs $(r^*, y^*)$ replace $(r, y)$ to guide the student model toward teacher-quality reasoning and answering.  

\begin{table*}[!t]
    \centering
    \tabcolsep 3pt
    \resizebox{\textwidth}{!}{%
    \begin{tabular}{lcccccccccccc}
    \toprule
    Model&   \multicolumn{11}{c}{\textbf{Test Dataset}} \\ 
    \midrule
    & \multicolumn{5}{c}{\textbf{MultiModalQA}} && \multicolumn{5}{c}{\textbf{WebQA}} \\ 
    \cline{2-6} \cline{8-12}
    & & \multicolumn{2}{c}{\textbf{EM}}  & \multicolumn{2}{c}{\textbf{F1}}     && & \multicolumn{2}{c}{\textbf{EM}}   & \multicolumn{2}{c}{\textbf{F1}}   \\
    \cline{2-6} \cline{8-12}

& FT (Real/Syn) & Real & Syn & Real & Syn     && FT (Real/Syn) & Real & Syn   & Real & Syn   \\
\cline{2-6} \cline{8-12}

    LLaVa-1.6-7B & 5k/5k & 64.61 & \textbf{69.68} & 73.13 & \textbf{76.52} &&  5k/5k &69.88 & \textbf{75.49}  & 78.27 & \textbf{87.61} \\
    LLaVa-1.6-7B & 10k/10k & 73.96 & \textbf{75.14} & 78.36 & \textbf{79.48} && 10k/10k & 77.49 & \textbf{79.06} & 82.59 & \textbf{82.76}\\
    LLaVa-1.6-7B & 23.8k/21k & 78.79 & \textbf{79.41} & 82.35 & 80.65 && 34.2k/16k & 81.36 & \textbf{82.48} & \textbf{85.79} & 82.83\\

    LLaVa-1.6-13B & 10k/10k & 77.45 & \textbf{79.46} & 79.12 & \textbf{81.32} && 10k/10k & 80.22 & \textbf{83.24} & 84.26 & \textbf{84.95}\\

    LLaVa-1.6-13B & 23.8k/21k & 82.95 & \textbf{83.56} & 83.71 & \textbf{84.76} && 34.2k/13k & 83.36 & \textbf{85.34} & 86.92 & \textbf{87.64} \\

    InternVL-2-8B & 5k/5k &  69.43 & \textbf{73.92} & 79.77 & \textbf{84.27} &&  5k/5k & 78.19 & \textbf{81.27} & 87.64 & \textbf{90.21}\\

    InternVL-2-8B & 10k/10k &  76.42 & \textbf{77.25} &  86.42 & \textbf{87.06} &&  10k/10k & 83.86 & \textbf{85.34}  &  91.82 & \textbf{94.05} \\

    InternVL-2-8B & 23.8k/17k &  81.36 & \textbf{82.95} &  \textbf{90.2} & 89.24  &&  34.2k/15k & 85.67 & \textbf{86.58}  &  88.05 & \textbf{92.23} \\

     InternVL-2-26B & 10k/10k & 78.79 & \textbf{79.23} & \textbf{88.91} & 88.44 && 10k/10k & 85.76 & \textbf{86.49} & 92.23 & \textbf{93.19}\\

     InternVL-2-26B & 23.8k/16k &  84.6 & \textbf{85.29} &  89.76 & \textbf{91.24} &&  34.2k/15k & 86.36 & \textbf{87.74}  &  \textbf{91.82} & 90.21 \\

     Idefics-2-8B & 5k/5k & 67.19 & \textbf{71.48} & 77.42& \textbf{79.32} && 5k/5k & 75.34& \textbf{78.31}& 84.51& \textbf{86.92} \\

      Idefics-2-8B & 10k/10k & 75.4 & \textbf{76.57} & \textbf{85.39}& 83.71&& 10k/10k & 80.11& \textbf{83.86} & 88.18 & \textbf{90.07}\\

       Idefics-2-8B & 23.8k/18k & 81.37 & \textbf{81.85} & 89.12 & \textbf{89.76} && 34.2k/15k & 87.49 & \textbf{87.67}& 91.82 & \textbf{92.23} \\

\midrule 
    
LLaVa-1.6-7B & None & \multicolumn{2}{c}{50.85} & \multicolumn{2}{c}{56.34} && None & \multicolumn{2}{c}{56.37} & \multicolumn{2}{c}{65.44}\\
    LLaVa-1.6-13B & None & \multicolumn{2}{c}{56.83} & \multicolumn{2}{c}{61.17} && None &\multicolumn{2}{c}{61.79} & \multicolumn{2}{c}{68.32}\\

    InternVL-2-8B & None & \multicolumn{2}{c}{61.27} & \multicolumn{2}{c}{68.26} && None & \multicolumn{2}{c}{68.01} & \multicolumn{2}{c}{76.39}\\

    InternVL-2-26B & None & \multicolumn{2}{c}{\textbf{68.39}} & \multicolumn{2}{c}{\textbf{74.08}} && None & \multicolumn{2}{c}{\textbf{74.59}} & \multicolumn{2}{c}{\textbf{80.61}}\\

    Idefics2-8B & None & \multicolumn{2}{c}{60.47} & \multicolumn{2}{c}{66.41} && None & \multicolumn{2}{c}{64.79} & \multicolumn{2}{c}{72.38}\\

\midrule 
    
GPT-4o & None & \multicolumn{2}{c}{\textbf{83.56}} & \multicolumn{2}{c}{\textbf{87.91}} && None & \multicolumn{2}{c}{\textbf{86.49}} & \multicolumn{2}{c}{\textbf{90.23}}\\

    Llama-3.2-90B & None & \multicolumn{2}{c}{77.18} & \multicolumn{2}{c}{80.37} && None &\multicolumn{2}{c}{82.22} & \multicolumn{2}{c}{86.9}\\

    Llama4-Scout & None & \multicolumn{2}{c}{79.13} & \multicolumn{2}{c}{83.68} && None & \multicolumn{2}{c}{84.46} & \multicolumn{2}{c}{88.72}\\

    Claude-3.5-Sonnet & None & \multicolumn{2}{c}{73.84} & \multicolumn{2}{c}{77.38} && None & \multicolumn{2}{c}{76.29} & \multicolumn{2}{c}{79.43}\\

    Claude-4-Sonnet & None & \multicolumn{2}{c}{78.25} & \multicolumn{2}{c}{82.92} && None & \multicolumn{2}{c}{82.18} & \multicolumn{2}{c}{87.61}\\
    
    \bottomrule
    \end{tabular}
    }
    \caption{Comparison of synthetic and human-annotated data across various models. We generated 18k synthetic training samples in total, but models are typically trained on 5k or 10k subsets for efficiency. Larger subsets (\(>10\)k) are only used in cases where we seek the minimum number of synthetic samples needed to outperform training on the full human-labeled set. For smaller models, we evaluate with 5k, 10k, and full training sets (23.8k for MultiModalQA, 34.2k for WebQA), while larger models use 10k and full sets. For each model, the listed synthetic sample size is the smallest (divisible by 1k) that surpasses the same model trained on the full human-labeled set. For real samples, we used the WebQA training set for testing on the WebQA test set, and similarly for MultiModalQA. “None” indicates default pretrained models. For extended results, see Appendix~\ref{appendix:more_models}.}
    \label{tab:main_results_formatted}
    \vspace{-0.9em}
\end{table*}

\subsection{Comparison with Human-Annotated Datasets}

Unlike prior methods like MultiModalQA \citep{talmor2021multimodalqa} and WebQA \citep{WebQA21}, our approach is fully automated with no human involvement (minimal human evaluation was used in creating the \benchmark\ only). This section compares the quality of our synthesized data against these human-annotated datasets. We trained LLaVA-1.6 \citep{liu2023llava, liu2023improvedllava, liu2024llavanext}, InternVL-2 \citep{chen2023internvl, chen2024far}, and Idefics-2 \citep{laurencon2023obelics, laurençon2024matters} on varying sizes of WebQA and MultiModalQA, evaluating on their respective test sets. We also trained the same models on \ourmethod-generated training data and evaluated them on the same test sets to assess the effectiveness of the synthesized data in comparison to human-annotated datasets.

As shown in Table \ref{tab:main_results_formatted}, models trained on \ourmethod\ data outperform those trained on original datasets, despite using longer documents. WebQA seems easier than MultiModalQA, with better performance from fewer samples. On average, EM improved by 1.81 for MultiModalQA and 1.96 for WebQA using equal or fewer synthetic samples. While EM gains often lead to higher F1, some cases show F1 drops due to hallucinated answers reducing string overlap. Models trained on fewer synthetic samples from \ourmethod\ match the performance of those trained on full datasets, showing faster convergence (see Section \ref{subsection:learning_efficiency}). Larger models also perform better with the same synthetic data; e.g., LLaVA-1.6-13B vs. 7B. GPT-4o leads among large LVLMs, likely due to its role in data generation (Refer to Appendix \ref{appendix:FT_performance} for qualitative analysis). 

When comparing these human annotated data with our synthesized data, a common question is: "Why not paraphrase existing datasets instead of synthesizing new ones?". The answer is paraphrasing does not enable domain adaptation, and as shown in Appendix \ref{appendix:paraphrasing}, fine-tuning on our synthetic data outperforms training on a paraphrased version of MultiModalQA.

\begin{figure*}[!t]
    \centering
    \includegraphics[width=\textwidth]{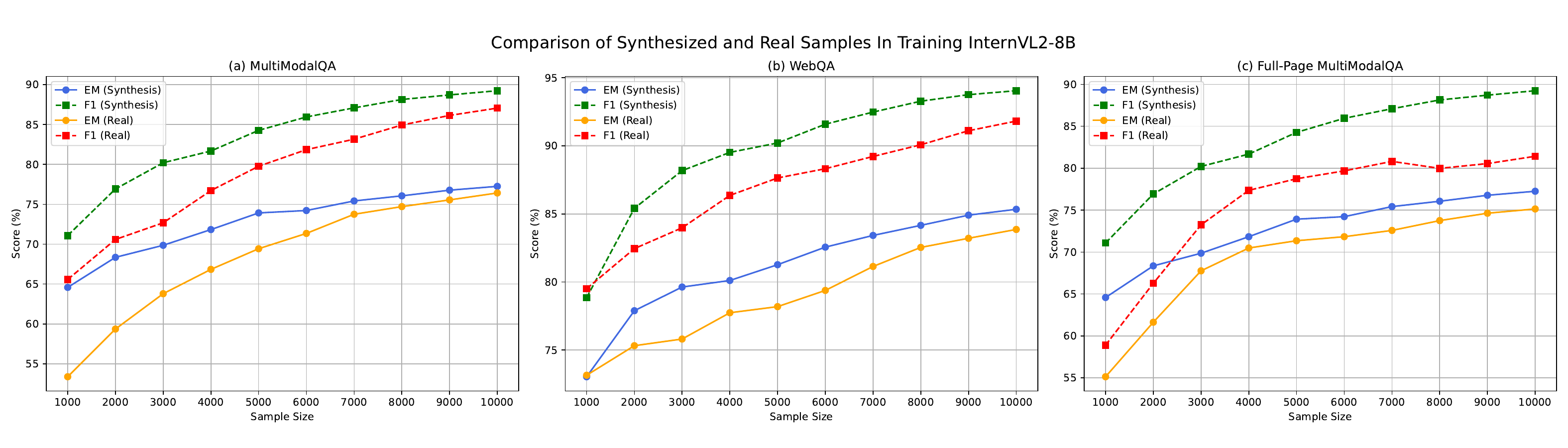}
    \caption{\textbf{\textit{(a) and (b)}}: EM and F1 comparison on 1k–10k samples for InternVL-2-8B shows that \ourmethod's synthetic data outperforms human-annotated data, with the gap narrowing as sample size increases. \textbf{\textit{(c)}}: Similar comparison using full Wikipedia pages from the MultimodalQA dataset to match our synthetic data format.}
    \label{fig:comparison_all}
\end{figure*}

\subsection{Learning Efficiency Comparison}
\label{subsection:learning_efficiency}

To evaluate learning efficiency, we ran experiments with InternVL-2-8B using incremental training sizes from 1k to 10k (in 1k steps) on both synthetic and human-annotated data. For real data, we used the same dataset for training and testing; e.g., when testing on WebQA, training samples were taken from the WebQA training set. 

As shown in Figure \hyperref[fig:comparison_all]{\ref*{fig:comparison_all}(a \& b)}, our synthesized data outperforms real data at smaller training sizes, though the gap narrows as sample size grows. Near 10k samples, learning efficiency with synthetic data declines more than with real data, likely due to its broader knowledge coverage. While this diversity aids early learning, it can lead to saturation, unlike real data, which offers more focused patterns and sustains steady learning \citep{hong-etal-2023-diminishing, maini-etal-2024-rephrasing}.

In a related experiment on MultiModalQA, we used full Wikipedia pages via linked articles as training data instead of information snippets. This was not possible for WebQA, as its source links mostly point to WikiMedia pages with limited text. Figure \hyperref[fig:comparison_all]{\ref*{fig:comparison_all}(c)} shows that models trained on full Wikipedia pages initially achieve better improvement per 1k samples; however, this advantage declines after approximately 3k samples. This suggests that, while full-page real data offers some early benefits over short snippets, it still lacks the generality and consistency of our synthesized dataset. In contrast, the synthesized data, with its built-in queries and diverse content, continues to support steady learning, likely due to its broader coverage and higher quality.

\subsection{Cross-Dataset Evaluation}
\label{subsec:cross_dataset}

To assess the generalizability of our synthesized data, we conducted a cross-dataset evaluation using InternVL-2-8B. We used 1k samples from \benchmark\ and 1k from the MultiModalQA test set, both with full Wikipedia pages as sources. The model was trained separately on 5k samples from our dataset and 5k from MultiModalQA, using the same input format. As shown in Table \ref{tab:cross_eval}, the model trained on our data outperformed the one trained on MultiModalQA across both test sets, demonstrating stronger generalization. This also reflects the greater complexity and diversity of our benchmark, compared to MultiModalQA's template-based questions. See Appendix \ref{appendix:generalizability} to see the performance on model trained on \ourmethod's data when using on a out of domain MMQA dataset.

\begin{table}[!h]
    \centering
    \resizebox{\linewidth}{!}{%
    \begin{tabular}{c|cccc}
    \hline
    \multirow{3}{*}{\textbf{Training Set}} & \multicolumn{4}{c}{\textbf{Test Set}}                                 \\ \cline{2-5} 
                                  & \multicolumn{2}{c|}{\textbf{MultiModalQA}} & \multicolumn{2}{c}{\textbf{Ours}} \\ \cline{2-5} 
                                  & EM   & \multicolumn{1}{c|}{F1}    & EM         & F1          \\ \hline
    \textbf{MultiModalQA}         & 63.2 & \multicolumn{1}{c|}{74.24} & 48.6       & 61.45       \\
    \textbf{Ours}                 & \textbf{68.4} & \multicolumn{1}{c|}{\textbf{77.88}} & \textbf{55.8}       & \textbf{69.06}       \\ \hline
    \end{tabular}
    }
    \caption{Cross-Dataset Evaluation Results With InternVL-2-8B for MultiModalQA and Our Synthesized Benchmark.}
    \label{tab:cross_eval}
    \vspace{-0.5 cm}
\end{table}

\subsection{Key Stages in \ourmethod}

\begin{table}
    \centering
    \resizebox{\linewidth}{!}{%
    \begin{tabular}{l|cccc}
    \hline
    \multirow{3}{*}{\textbf{Steps Included}} & \multicolumn{4}{c}{\textbf{Test Dataset}}                                       \\ \cline{2-5} 
                                             & \multicolumn{2}{c|}{\textbf{MultiModalQA}} & \multicolumn{2}{c}{\textbf{WebQA}} \\ \cline{2-5} 
                                             & EM        & \multicolumn{1}{c|}{F1}        & EM               & F1              \\ \hline
    \textbf{-}                               & 62.78     & \multicolumn{1}{c|}{64.39}     & 69.32            & 78.56           \\
    Question Validation                            & 64.61     & \multicolumn{1}{c|}{70.12}     & 70.28            & 78.44           \\
    + Answer Validation                       & 67.96     & \multicolumn{1}{c|}{73.56}     & 74.59            & 79.67           \\
    + Query                 & 70.83     & \multicolumn{1}{c|}{75.84}     & 77.49            & 81.86           \\
    + Query Validation             & \textbf{73.92}     & \multicolumn{1}{c|}{\textbf{84.27}}     & \textbf{81.27}            & \textbf{90.21}           \\ 
    \hline
    \end{tabular}
    }
    \caption{Results of the \ourmethod\ with and without key steps like query generation and verification. All other steps are included in all of the results. The plus sign ("+") at the start each steps means all the previous steps were included as well.}
    \label{tab:steps_effect}
\end{table}

To assess the impact of each data filtering and validation step, we evaluated InternVL-2-8B on 5k training samples under different configurations. As shown in Table \ref{tab:steps_effect}, each step contributed to performance gains. Question validation improved relevance by filtering questions tailored to the complex, multimodal, multihop requirements of the test sets, and reduced hallucinations, boosting F1 with more accurate answers. Answer validation removed incorrect samples, while query generation distilled knowledge from larger models, improving EM and F1. Query validation reinforced consistency by ensuring proper structure. Refer to Appendix \ref{appendix:each_stage_stat} for statistics like \textbf{"Rejection Rate"} on the validation stages used to remove specific samples during data synthesis.

\subsection{\benchmark\ Evaluation Results}

\begin{figure}
    \centering
    \includegraphics[width=\linewidth]{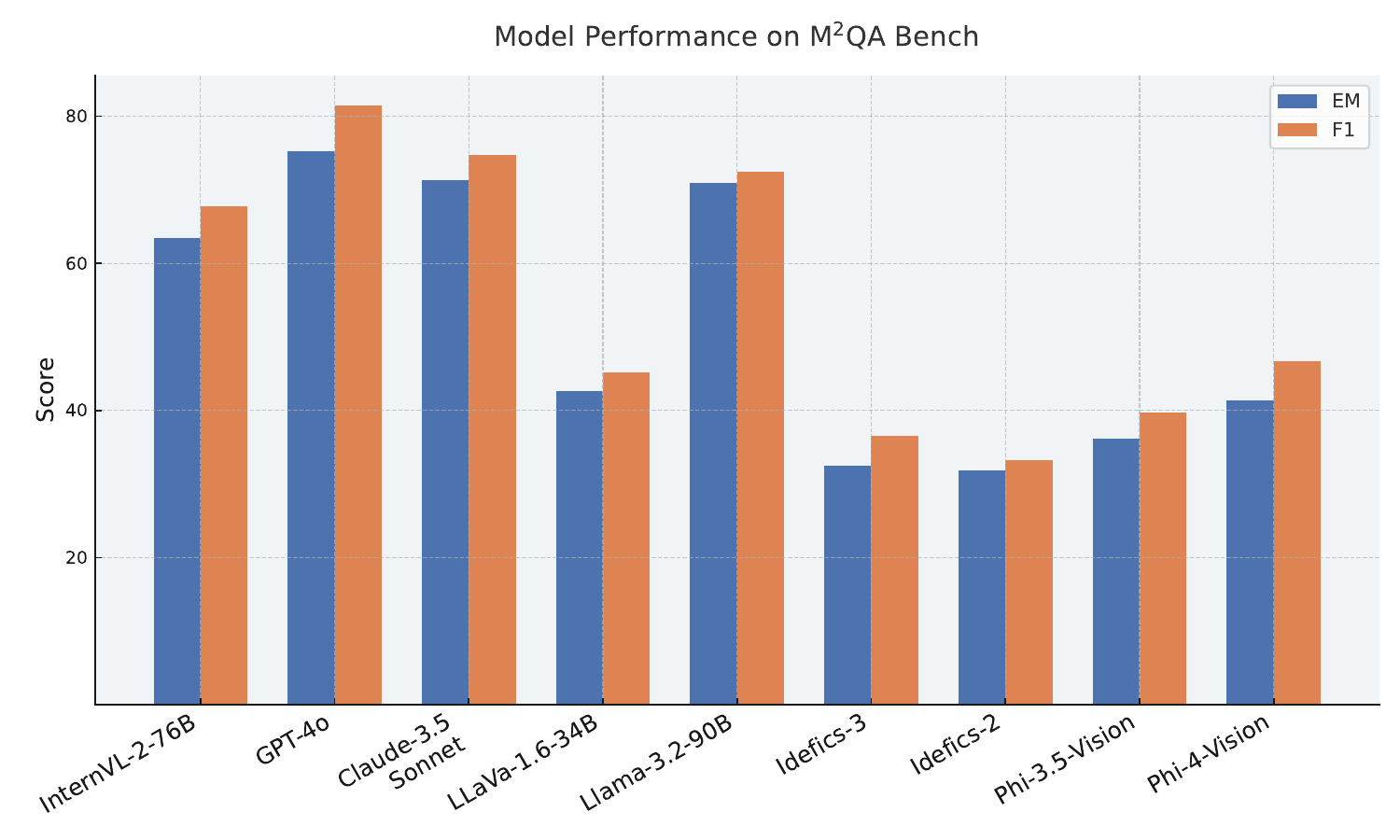}
    \caption{
        Bar plot showing EM and F1 scores of multimodal models on \benchmark.
    }
    \label{fig:bench_results}
\end{figure}

To assess the complexity of \benchmark\ and compare model performance, we evaluated a diverse set of models, as shown in Figure \ref{fig:bench_results}. 
GPT-4o stands out, outperforming even larger models like LLaMa-3.2-90B and Claude-3.5-Sonnet by a notable margin. 
Interestingly, smaller models in the 4B–8B range, particularly those from the Phi family \citep{abdin2024phi, abdin2024phi4}, also achieve competitive results despite their scale. 
GPT-4o’s advantage may stem in part from its involvement in the initial data generation, but this remains an open question, as human annotators thoroughly reviewed and corrected the dataset to ensure fairness, reduce bias, and maintain high quality.

\section{Human Evaluation of Answer Validation}
\label{sec:human_eval}

\begin{figure}
    \centering
    \includegraphics[width=\linewidth]{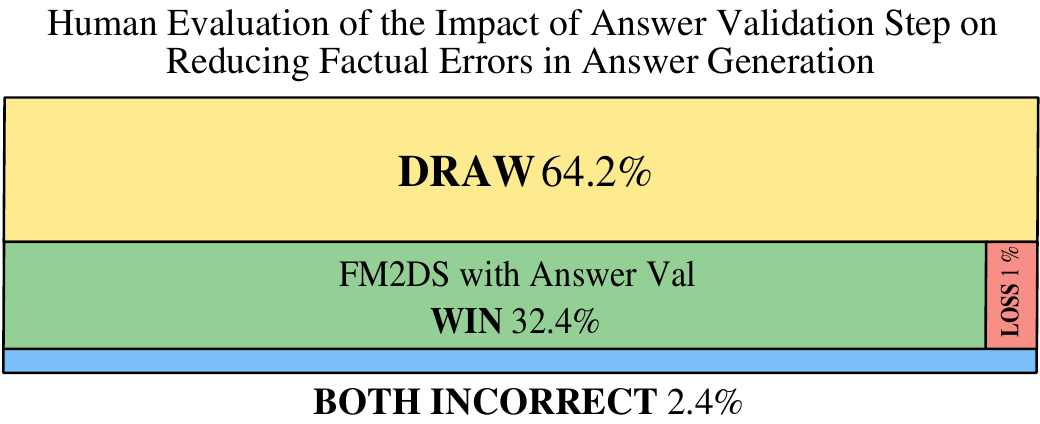}
    \caption{Human evaluation results, more people preferred \ourmethod\ with answer validation (green region) than without (red region).}
    \label{fig:human_eval}
\end{figure}

To evaluate the accuracy and impact of our automatic answer validation component, we conducted a human evaluation study. 
After generating questions for 100 samples, we continued the pipeline using two methods: \ourmethod\ with and without answer validation. 
These samples were divided into four batches of 25, with each batch evaluated by three participants to mitigate potential bias or human errors. 
Twelve participants used our evaluation platform (Appendix~\ref{appendix:human_evaluation_details}) to judge answer correctness.  

Participants were instructed to verify the correctness of each answer by reviewing the associated Wikipedia pages. Once they determined the accuracy of each answer, they were asked to select one of four options: (1) Method 1's answer is correct, (2) Method 2's answer is correct, (3) both methods generated the correct answer, or (4) neither answer is correct (Method 1 and Method 2 were randomly assigned to the datasets generated with and without answer validation, respectively.). 
The results, shown in Figure \ref{fig:human_eval}, reveal a clear trend: answer validation increases the likelihood of correct answers, even though the model without validation still produced correct answers in over 60\% of cases. 
The \textbf{Fleiss’ Kappa} was \textbf{0.91}, indicating strong agreement, though this is expected, as the task involved factual questions with definitive answers rather than subjective judgments.

\section{Conclusion \& Future Works}

We present \ourmethod, a novel methodology for synthesizing high-quality data for multimodal multihop question answering. Unlike existing approaches that are limited to single-hop and single-modality settings, \ourmethod\ generates complex QA pairs that require reasoning over multiple modalities and sources, with minimal human intervention. Our framework enables the creation of a large-scale dataset that significantly boosts model performance, surpassing models trained on human-curated data in terms of test accuracy. These results demonstrate the effectiveness of synthetic data for advancing the state of the art in multimodal multihop QA. Additionally, \ourmethod\ offers a scalable and efficient solution for training data-hungry language models. For future work, we plan to synthesize MMQA samples using sources beyond Wikipedia, including multilingual content, code snippets, videos, and other diverse information type.

\clearpage

\section*{Limitations}

While \ourmethod\ offers a robust pipeline for synthesizing high-quality multimodal multihop QA data, several limitations remain.

First, although the framework incorporates strong validation steps, including factual consistency checks, named entity alignment, and hallucination detection, it is not immune to errors. Subtle factual inaccuracies and hallucinations may still persist, especially in answers grounded in complex visual content. Despite using multiple generations and automated agreement checks, there is still a risk that some incorrect samples pass through undetected.

Second, our reliance on large-scale generative models such as GPT-4o throughout multiple stages, including question generation, answer synthesis, captioning, and validation, makes the pipeline computationally expensive. This cost is further amplified by the need to regenerate failed samples that do not pass intermediate validation steps. In some settings, particularly when generating large-scale datasets, the repeated use of high-capacity models may pose practical limitations in terms of both time and resources.

Third, while our method improves factual accuracy and reduces hallucination, the validation pipeline is primarily designed for fact-based QA. This makes it less suitable for tasks involving subjective reasoning, commonsense inference, or open-ended discussion questions. Extending the pipeline to handle such cases would require fundamentally different validation strategies that go beyond factual grounding.

Finally, since the student models are trained on data generated by large models (used in both synthesis and supervision), there is a risk of knowledge leakage or model bias propagation. The synthetic data may overrepresent patterns and linguistic preferences from the teacher models, potentially limiting the generalizability of the student models trained on it.

\section*{Ethics Statement}
\paragraph{Potential Risks}
The primary risk associated with this work lies in the possibility of propagating factual inaccuracies or biases through automatically synthesized data. While our validation pipeline aims to minimize hallucinations and ensure factual correctness, it may not catch all subtle errors. Additionally, overreliance on large language models for data generation could inadvertently reinforce biases encoded in those models. Our approach does not involve any sensitive personal data or downstream applications that could directly harm individuals.

\paragraph{Annotator Recruitment}
To verify and refine the samples in our \benchmark\ dataset, we recruited three human annotators with prior experience in NLP and data annotation (two men and one woman). These annotators were compensated fairly at a rate of \$25 per hour to reflect their expertise and time investment. All annotators were provided with detailed task descriptions and underwent an informed consent process prior to participation. The annotation process was conducted in accordance with ethical research guidelines and ensured voluntary participation and data confidentiality.

\paragraph{Evaluator Recruitment}
For the human evaluation component of our study, we recruited twelve evaluators to compare answers generated with and without our validation pipeline. Evaluators were compensated at a rate of \$10 per hour and participated voluntarily after giving informed consent. They were clearly informed about the nature and purpose of the task. We ensured the task was low-risk, did not involve sensitive content, and that participation remained anonymous and non-intrusive.

\paragraph{Consent and Data Privacy}
All participants in both annotation and evaluation roles were briefed on the nature of the task and explicitly consented to take part in the study. No personally identifiable information was collected or stored during any part of the research process. All data generated and reviewed by annotators and evaluators remained anonymous and was used strictly for academic research purposes.

\paragraph{Use of AI Assistants}
We used AI assistants such as GitHub Copilot and ChatGPT to support coding, text editing, and formatting tasks during the development of this paper and the implementation of our framework. These tools were employed to accelerate workflow and refine writing, but all conceptual, experimental, and analytical decisions were made by the authors. We ensured that no sensitive data was provided to these tools during usage.

\clearpage

\bibliography{custom}

\appendix

\clearpage

\section{Prompts}
\label{appendix:prompts}

In our data generation pipeline, \ourmethod, which incorporates LVLMs, we carefully designed prompts to guide the model through tasks involving cross-modal reasoning and data synthesis. Each prompt was carefully designed with specific elements to ensure precision, clarity, and completeness in achieving the task's objectives, while also minimizing the need for error correction during the evaluation process. In the following sections, we outline the rationale behind the structure and components of these prompts.

\begin{prompt}[!ht]
    \centering
    \begin{promptbox}{Question Generation Prompt}
        \texttt{
            Generate a multi-hop question based on the provided information. A multi-hop question requires the model to utilize information from all available documents in combination to reach the correct answer. Specifically, the question should be designed to be unanswerable if any one of the documents is missing. Furthermore, focus on creating questions that compel the model to extract and synthesize relevant information across multiple modalities—such as images and text. This means that answering the question correctly will demand integrating insights from each source and modality, making it impossible to arrive at an accurate answer using any single document or modality alone.
        }
        
        \vspace{0.4cm}
        
        \texttt{Here are the documents:}
        
        \texttt{[Documents]}
    
        \vspace{0.4cm}
    
        \texttt{Here are examples:}
    
        \texttt{[Example(s)]}
    \end{promptbox}
    \caption{The prompt for question generation defines what constitutes a multi-hop question and instructs the model to create a multimodal and multihop question based on the provided documents. It emphasizes that the question should require information from multiple modalities and multiple given documents to be answered, similar to the given example(s).}
    \label{prompt:question}
\end{prompt}

\begin{prompt}[!ht]
    \centering
    \begin{promptbox}{Answer Generation Prompt}
    \texttt{
        You are provided with multiple documents, including both textual content and images, along with a question. Your task is to carefully review each document, analyze the images, and derive an answer based on the information contained across all sources. Aim to combine insights from both documents and across modalities to deliver the most accurate and comprehensive response possible.
    }

    \vspace{0.4cm}

    \texttt{Question:}
    
    \texttt{[Question]}

    \vspace{0.4cm}
    
    \texttt{Here are the documents:}
    
    \texttt{[Documents]}

    \vspace{0.4cm}

    \texttt{Here are examples:}

    \texttt{[Example(s)]}

    \end{promptbox}
    \caption{The answer generation prompt instructs the model to answer the question solely based on the provided documents, utilizing all available modalities, without relying on its pre-trained knowledge.}
    \label{prompt:answer}
\end{prompt}

\subsection{Question Generation}

Prompt \ref{prompt:question} show the prompt used for question generation. Using this prompt, we ask the model to create multi-hop questions that require information from all provided documents and modalities (e.g., text and images) to answer. The key aim is to design questions that are unanswerable if any one document or one modality is given, promoting the need for multi-document and multimodal reasoning. It ensures the model generates questions that require synthesizing information from diverse sources to form a comprehensive understanding. To avoid duplicate data generation, if the generated question was already present in the dataset, we reused the same prompt but included the previously generated questions from the same set of documents. The model was then instructed to generate a new, unique question.

\subsection{Answer Generation}

The prompt for answer generation directs the model to analyze multiple documents, encompassing both text and images, to address the given question. It emphasizes integrating and synthesizing information from all sources to deliver the most accurate and comprehensive response. The prompt ensures that the model considers all modalities and documents without relying solely on a single source or the model's pre-trained knowledge, focusing exclusively on the provided materials. Refer to Prompt \ref{prompt:answer} for the answer-generation prompt.

\subsection{Query Generation}

As illustrated in Prompt \ref{prompt:query}, in query generation, the model is tasked with explaining the step-by-step process used to extract relevant information from the documents and determine the answer based on the extracted snippets. This task emphasizes transparency by requiring the model to identify the relevant sections of each document and describe how information from multiple sources is retrieved and combined to arrive at the correct answer, promoting explainability in the model's reasoning process.

\begin{prompt}[!ht]
    \centering
    \begin{promptbox}{Query Generation Prompt}
        \texttt{
            You are provided with multiple documents, a question, and the answer. Your task is to explain the step-by-step process you would use to extract and verify the answer using information from the documents and various modalities. Clearly identify each document title and relevant sections, and describe how you locate, interpret, and integrate information across both documents to derive the correct answer.
        }
    
        \vspace{0.4cm}
    
        \texttt{Question:}
        
        \texttt{[Question]}
    
        \vspace{0.4cm}
    
        \texttt{Answer:}
        
        \texttt{[Answer]}
    
        \vspace{0.4cm}
        
        \texttt{Here are the documents:}
        
        \texttt{[Documents]}
    
    \end{promptbox}
    \caption{The query generation prompt instructs the model to provide a step-by-step plan for extracting relevant information needed to answer the question.}
    \label{prompt:query}
\end{prompt}

\section{Experimental Settings}
\label{appendix:exp_settings}

In this work, we conducted experiments on a cluster of \textbf{8 NVIDIA H100 80GB GPUs}. The distributed setup allowed us to efficiently scale our fine-tuning process across multiple devices. The fine-tuning process was carried out using low-rank adaptation (LoRA) \citep{hu2022lora}, a technique for efficient adaptation of pretrained models with low-rank matrices, reducing the number of trainable parameters. The key hyperparameters used in the fine-tuning procedure include a learning rate of 1e-4, a batch size of 8 per device (totaling 64 across 8 devices), LoRA rank set to 8, LoRA alpha set to 32, a weight decay of 0.01, and the number of epochs was 5. Additionally, AdamW optimizer was used with $\beta_1 = 0.9$, $\beta_2 = 0.98$, and $\epsilon = 1e-8$. The models were fine-tuned using mixed-precision training to take full advantage of the 80GB memory on each H100 GPU.  For inference time, we set the temperature to 0.7, which strikes a balance between randomness and coherence in the model's responses, producing more varied outputs without sacrificing too much quality. This setup ensured efficient usage of computational resources while maintaining high model performance.

\section{Cost of Sample Generation}
\label{appendix:sample_cost}

We analyze the cost of generating high-quality samples across different approaches. Using GPT-4o, the average cost of producing one high-quality sample is approximately \$0.035. By comparison, LLaMA~3.2-90B achieves similar quality at the cost of 42 H100 GPU hours for generating 5k samples (see Table \ref{tab:model_comparison_data_synthesis} in Appnedix \ref{appendix:other_generative_models}), which is competitive with unimodal data synthesis methods. This aligns with prior work on unimodal data generation \cite{chen-etal-2024-shot}. 

In contrast, human-written samples such as those in MMQA are substantially more expensive: each sample costs around \$2 and requires approximately 5 minutes of human effort. This comparison highlights the scalability and cost-effectiveness of large language models for multimodal data synthesis, offering orders-of-magnitude savings over manual annotation while maintaining comparable quality.

\section{The Effect of the Number of In-Context Examples}
\label{appendix:different_shots}

\begin{table}[!t]
    \centering
    \resizebox{\linewidth}{!}{%
    \begin{tabular}{c|cc|cc}
    \hline
    \multirow{2}{*}{\textbf{Number of Shots}} & \multicolumn{2}{c|}{\textbf{MultiModalQA}} & \multicolumn{2}{c}{\textbf{WebQA}} \\
     & EM & F1 & EM & F1 \\
    \hline
    \textbf{zero-shot} & 66.42 & 75.1 & 71.05 & 79.3 \\
    \textbf{one-shot} & 73.92 & 84.27 & 81.27 & 90.21 \\
    \textbf{two-shot} & 74.3 & 83.71 & 81.5 & \underline{91.37} \\
    \textbf{three-shot} & \underline{74.45} & \underline{85.39} & \underline{81.6} & 91.21 \\
    \hline
    \end{tabular}
    }
    \caption{Effect of the number of in-context documents on the performance of Intervl-2-8B on MultiModalQA and WebQA datasets.}
    \label{tab:in_context_effect}
\end{table}

Table \ref{tab:in_context_effect} presents the results of evaluating the Intervl-2-8B model with varying numbers of in-context examples on the MultiModalQA and WebQA datasets. In the zero-shot setting, \ourmethod\ exhibits limited understanding of multimodal multi-hop question answering, and occasionally circumvents the validation step by simply generating a question that is not multihop. For example, "\textit{Looking at the image of the Eiffel Tower, what engineering innovation allowed it to surpass previous structures in height?}" prompts the model to use the image, but the answer is available in the page's text on tall structures. As we move from zero-shot to one-shot, there is a significant boost in EM and F1 scores, reflecting improved performance with minimal context. The improvement from one-shot to two-shot is marginal, suggesting diminishing returns. With three in-context samples, the gains become minimal, indicating that additional samples beyond two provide little benefit. This diminishing return may stem from the model's limited context window, which restricts its ability to fully utilize large in-context samples \citep{kaplan2020scaling, bertsch2024context}.

\section{Impact of Data Synthesis LLM Choice}
\label{appendix:other_generative_models}

\begin{table}[h!]
    \centering
    \resizebox{\linewidth}{!}{%
    \begin{tabular}{c|cc|cc}
    \hline
    \multirow{2}{*}{\textbf{Model}} & \multicolumn{2}{c|}{\textbf{MultiModalQA}} & \multicolumn{2}{c}{\textbf{WebQA}} \\
    & EM & F1 & EM & F1 \\
    \hline
    \textbf{GPT-4o} & \underline{73.92} & \underline{84.27} & \underline{81.27} & \underline{90.21} \\
    \textbf{Claude-3.5-Sonnet} & 68.35 & 76.01 & 76.98 & 83.86 \\
    \textbf{Llama-3.2-90B} & 70.64 & 76.32 & 79.81 & 86.75 \\
    \hline
    \end{tabular}
    }
    \caption{Performance comparison of different models for data generation on test datasets.}
    \label{tab:model_comparison_data_synthesis}
\end{table}

\begin{table}[!h]
    \centering
    \resizebox{\linewidth}{!}{%
    \begin{tabular}{l|c|ccc}
    \hline
    \textbf{Model} & \textbf{Setting} & \textbf{MultimodalQA} & \textbf{WebQA} & \textbf{\benchmark} \\
    \hline
    \textbf{LLaVA-1.6-13B} & FT  & -17.4 & -14.2 & -18.6 \\
    \textbf{LLaVA-1.6-13B} & PT  & -22.7 & -19.6 & -24.1 \\ \hline
    \textbf{InternVL-2-8B} & FT  & -19.1 & -16.3 & -20.2 \\
    \textbf{InternVL-2-8B} & PT  & -25.3 & -21.0 & -26.8 \\ \hline
    \textbf{GPT-4o}        & PT  & -15.9 & -13.1 & -17.3 \\
    \hline
    \end{tabular}
    }
    \caption{Exact Match changes ($\Delta$EM) is measured when models are prompted without supporting context. We report the change as $\Delta$EM on MultimodalQA, WebQA, and \benchmark. Here, FT denotes fine-tuned models, while PT refers to pretrained-only models.}
    \label{tab:zeroshot}
\end{table}

\begin{table*}[!h]
    \centering
    \tabcolsep 3pt
    \resizebox{0.98\textwidth}{!}{%
    \begin{tabular}{lcccccccccccc}
        \toprule
        Model&   \multicolumn{11}{c}{\textbf{Test Dataset}} \\ 
        \midrule
        & \multicolumn{5}{c}{\textbf{MultiModalQA}} && \multicolumn{5}{c}{\textbf{WebQA}} \\ 
        \cline{2-6} \cline{8-12}
        & & \multicolumn{2}{c}{\textbf{EM}}  & \multicolumn{2}{c}{\textbf{F1}}     && & \multicolumn{2}{c}{\textbf{EM}}   & \multicolumn{2}{c}{\textbf{F1}}   \\
        \cline{2-6} \cline{8-12}
    
        & FT (Real/Syn) & Real & Syn & Real & Syn     && FT (Real/Syn) & Real & Syn   & Real & Syn   \\
        \cline{2-6} \cline{8-12}
    
        LLaVa-1.6-34B & 10k/10k & 79.41 & \textbf{80.92} & 82.55 & \textbf{83.71} && 10k/10k & 84.41 & \textbf{84.94} & 85.58 & \textbf{85.79} \\
        LLaVa-1.6-34B & 23.8k/16k & 84.83 & \textbf{85.29} & 85.51 & \textbf{86.42} && 34.2k/13k & 86.48 & \textbf{87.49} & \textbf{88.18} & \textbf{88.18} \\
        InternVL-2-40B & 10k/10k & 82.18 & \textbf{83.56} & 89.24 & \textbf{90.2} && 10k/10k & 87.67 & \textbf{89.77} & 92.76 & \textbf{93.82} \\
        InternVL-2-40B & 23.8k/15k & 86.63 & \textbf{87.27} & \textbf{91.73} & 91.44 && 34.2k/14k & 89.77 & \textbf{90.32} & \textbf{93.19} & \textbf{93.19} \\
        InternVL-2-76B & 10k/10k & 83.95 & \textbf{86.15} & 89.76 & \textbf{90.72} && 10k/10k & 88.12 & \textbf{90.32} & 93.19 & \textbf{94.77} \\
        InternVL-2-76B & 23.8k/14k & 90.82 & \textbf{91.34} & 92.79 & \textbf{93.81} && 34.2k/14k & 93.14 & \textbf{93.65} & \textbf{94.06} & 93.82 \\
        InternVL-2.5-8B & 5k/5k & 72.13 & \textbf{76.82} & 80.46 & \textbf{85.91} && 5k/5k & 79.27 & \textbf{84.74} & 86.39 & \textbf{91.18} \\
        InternVL-2.5-8B & 10k/10k & 75.41 & \textbf{78.92} & 83.73 & \textbf{88.16} && 10k/10k & 82.34 & \textbf{86.95} & 89.57 & \textbf{93.23} \\
        InternVL-2.5-26B & 5k/5k & 76.98 & \textbf{81.27} & 85.15 & \textbf{90.32} && 5k/5k & 85.62 & \textbf{89.79} & 91.37 & \textbf{94.56} \\
        InternVL-2.5-26B & 10k/10k & 80.42 & \textbf{83.84} & 88.29 & \textbf{92.17} && 10k/10k & 88.48 & \textbf{91.53} & 93.68 & \textbf{95.83} \\
        InternVL-3-9B & 5k/5k & 74.72 & \textbf{79.45} & 82.63 & \textbf{88.03} && 5k/5k & 83.19 & \textbf{87.57} & 89.42 & \textbf{93.14} \\
        InternVL-3-9B & 10k/10k & 77.36 & \textbf{81.71} & 85.18 & \textbf{89.94} && 10k/10k & 86.27 & \textbf{89.62} & 92.18 & \textbf{95.28} \\
        InternVL-3-38B & 5k/5k & 82.64 & \textbf{86.28} & 90.73 & \textbf{94.34} && 5k/5k & 89.94 & \textbf{91.93} & 94.27 & \textbf{96.35} \\
        InternVL-3-38B & 10k/10k & 85.18 & \textbf{88.31} & 92.86 & \textbf{95.73} && 10k/10k & 91.16 & \textbf{93.83} & 95.62 & \textbf{97.14} \\
        Idefics-3-8B & 10k/10k & 76.42 & \textbf{77.56} & 86.74 & \textbf{88.23} && 10k/10k & 82.48 & \textbf{84.62} & 89.12 & \textbf{90.09} \\
        Idefics-3-8B & 23.8k/19k & 82.18 & \textbf{83.56} & 90.27 & \textbf{91.81} && 34.2k/15k & 86.49 & \textbf{87.77} & 92.55 & \textbf{93.31} \\
        Phi-3.5-Vision-Instruct-4.2B & 10k/10k & 69.43 & \textbf{70.25} & 75.35 & \textbf{77.57} && 10k/10k & 78.31 & \textbf{80.22} & 84.5 & \textbf{85.18} \\
        Phi-3.5-Vision-Instruct-4.2B & 23.8k/22k & 77.85 & \textbf{78.79} & 82.59 & \textbf{84.61} && 34.2k/19k & 80.11 & \textbf{81.27} & 85.34 & \textbf{87.48} \\
        mPLUG-DocOwl-1.5-8B & 10k/10k & 72.24 & \textbf{74.82} & 78.82 & \textbf{79.49} && 10k/10k & 81.27 & \textbf{83.86} & 87.52 & \textbf{88.75} \\
        mPLUG-DocOwl-1.5-8B & 23.8k/20k & 79.41 & \textbf{80.12} & 84.69 & \textbf{87.07} && 34.2k/17k & 82.48 & \textbf{84.42} & 87.93 & \textbf{89.21} \\
          
        \midrule 
        
        LLaVa-1.6-34B & None & \multicolumn{2}{c}{60.11} & \multicolumn{2}{c}{64.06} && None & \multicolumn{2}{c}{64.33} & \multicolumn{2}{c}{70.82}\\
        InternVL-2-40B & None & \multicolumn{2}{c}{72.93} & \multicolumn{2}{c}{77.42} && None & \multicolumn{2}{c}{76.98} & \multicolumn{2}{c}{82.33}\\
        InternVL-2-76B & None & \multicolumn{2}{c}{\textbf{75.32}} & \multicolumn{2}{c}{79.32} && None & \multicolumn{2}{c}{78.31} & \multicolumn{2}{c}{85.41}\\
        InternVL-2.5-8B & None & \multicolumn{2}{c}{63.27} & \multicolumn{2}{c}{70.16} && None & \multicolumn{2}{c}{69.38} & \multicolumn{2}{c}{77.42} \\
        InternVL-2.5-26B & None & \multicolumn{2}{c}{70.35} & \multicolumn{2}{c}{76.82} && None & \multicolumn{2}{c}{76.18} & \multicolumn{2}{c}{83.28} \\
        InternVL-3-9B & None & \multicolumn{2}{c}{65.47} & \multicolumn{2}{c}{71.92} && None & \multicolumn{2}{c}{72.58} & \multicolumn{2}{c}{80.27} \\
        InternVL-3-38B & None & \multicolumn{2}{c}{75.13} & \multicolumn{2}{c}{\textbf{81.35}} && None & \multicolumn{2}{c}{\textbf{80.64}} & \multicolumn{2}{c}{\textbf{87.46}} \\
        Idefics-3-8B & None & \multicolumn{2}{c}{61.27} & \multicolumn{2}{c}{69.74} && None & \multicolumn{2}{c}{69.88} & \multicolumn{2}{c}{76.39}\\
        Phi-3.5-Vision-Instruct-4.2B & None & \multicolumn{2}{c}{55.78} & \multicolumn{2}{c}{62.16} && None & \multicolumn{2}{c}{63.68} & \multicolumn{2}{c}{69.74}\\
        mPLUG-DocOwl-1.5-8B & None & \multicolumn{2}{c}{58.46} & \multicolumn{2}{c}{64.02} && None & \multicolumn{2}{c}{66.38} & \multicolumn{2}{c}{71.26}\\
        \bottomrule
    \end{tabular}
    }
    \caption{Comparison of model performance across various architectures, sizes, and sample sources (real vs. synthesized by \ourmethod). The models were evaluated on 10k samples and the full dataset (23.8k samples for MultiModalQA and 34.2k samples for WebQA). When comparing models tuned on synthesized data with those trained on the full training set, the smallest number of synthetic samples (divisible by 1k) that outperforms models trained on the full datasets is reported. For real sample evaluations, the WebQA training set is used for testing on the WebQA test set, and the same applies to MultiModalQA. Models trained with synthesized samples consistently outperform those trained with equivalent numbers of real samples.}
    \label{tab:more_models}
\end{table*}

To evaluate the effectiveness of different methods for synthetic data generation, we compared three prominent language models: GPT-4o, Claude 3.5 Sonnet, and Llama-3.2-90B, as shown in Table \ref{tab:model_comparison_data_synthesis}. Using Intervl-2-8B with 5K fine-tuning samples as our baseline model, we tested the quality of generated data on two distinct datasets: MultiModalQA and WebQA. The results, measured using EM and F1 scores, demonstrate that GPT-4o consistently outperforms other models across both datasets. We also see that Llama-3.2-90B shows competitive performance as an open-source model with less number of parameters, particularly in WebQA tasks. Claude 3.5 Sonnet generally yields lower scores across both datasets. As shown in Figure \ref{fig:bench_results}, Claude-3.5-Sonnet outperforms Llama-3.2-90B, which may be attributed to differences in the tasks that were included during their respective training phases. This observation can be further investigated.

\begin{figure*}[!h]
    \centering
    \includegraphics[width=\textwidth]{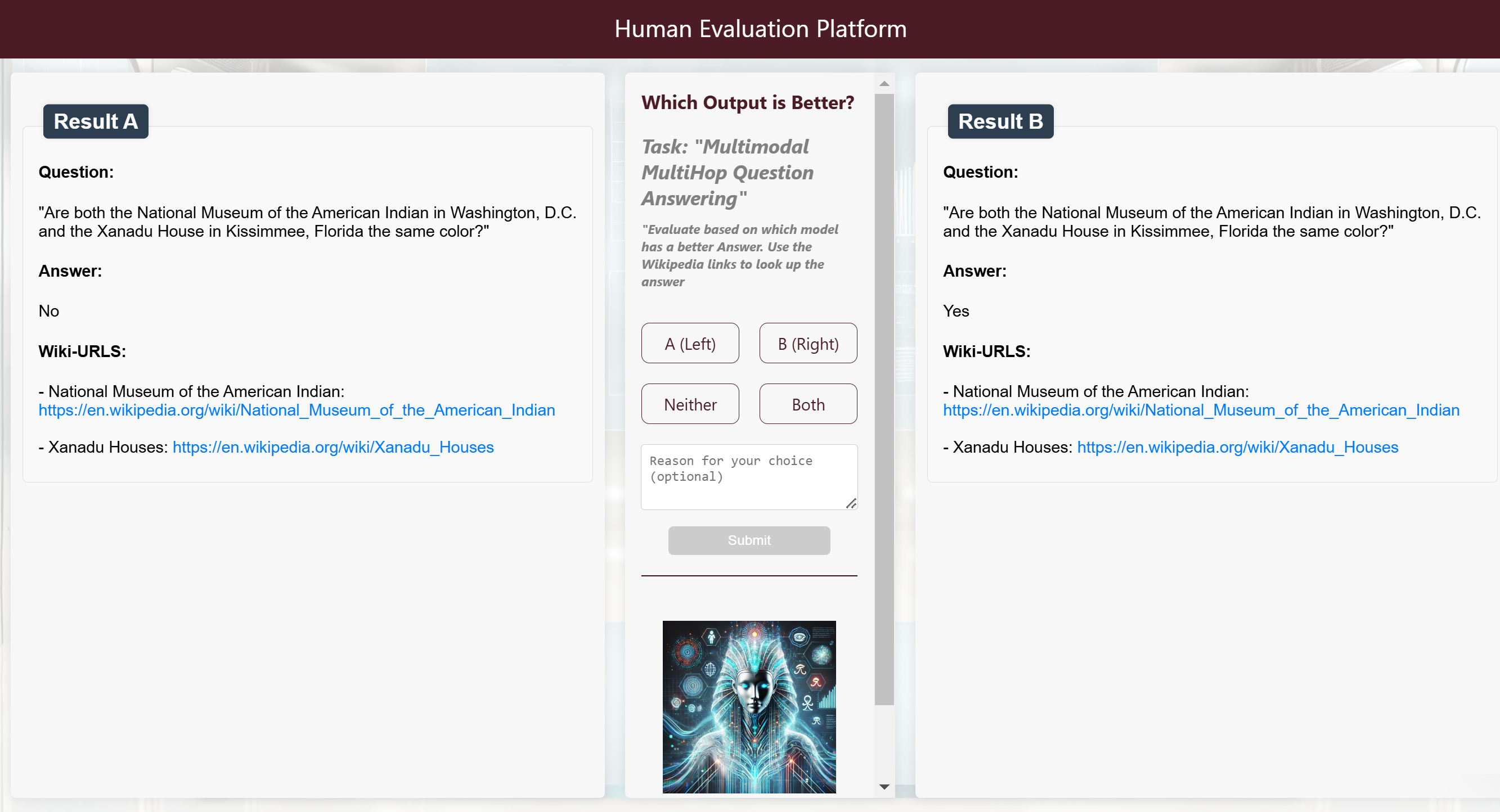}
    \caption{The custom evaluation application was used for human evaluation. The application presents each participant with a randomly selected question, relevant Wikipedia pages, and two model-generated answers labelled as \textit{Answer A} and \textit{Answer B}. One answer is generated by the pipeline with validation, while the other comes from the pipeline without it. Participants are asked to choose the correct answer and optionally provide feedback on their choice. To minimize bias, the application randomizes the position of each model's answer.}
    \label{fig:eval_app}
\end{figure*}

\section{Role of Supporting Context}
\label{appendix:zero_context_baseline}

An important question in multimodal multihop QA is whether large language models can answer questions directly in a zero-shot setting without access to supporting context. To investigate this, we evaluated a zero-shot baseline where models were only prompted with the question and no additional multimodal context. As shown in Table~\ref{tab:zeroshot}, performance dropped significantly across models compared to settings where full context was provided. 

Pre-trained models were the most affected by missing context. Fine-tuned models experienced smaller declines, indicating that \ourmethod\ training equips them with reasoning strategies that generalize beyond explicit evidence. Notably, GPT-4o, despite not being fine-tuned on \ourmethod, showed the smallest losses. This highlights GPT-4o’s strong inherent reasoning abilities, while also reinforcing that fine-tuning on \ourmethod\ fosters similar robustness even when no supporting context is available.

Moreover, the drop is consistently larger on \benchmark\, highlighting that its tasks depend more heavily on fine-grained contextual grounding. This underscores both the higher complexity of \benchmark\ and the central role of \ourmethod\ in training models that remain resilient even when explicit multimodal evidence is absent.

\section{Investigating Performance of More Models on \ourmethod\ Sythesized Data}
\label{appendix:more_models}

In addition to the models discussed in Section \ref{sec:results}, we explored other model families, including Idefics3 \citep{laurençon2024building}, mPLUG-DocOwl-1.5 \citep{hu2024mplugdocowl15unifiedstructure}, and Phi-3.5-Vision-Instruct \citep{abdin2024phi}, as well as larger versions within the explored families presented in Table \ref{tab:main_results_formatted}. The results in Table \ref{tab:more_models} demonstrate the reliability of our data synthesis approach, which consistently enhances model performance across all models and sizes compared to an equivalent number of real samples. 

As Table \ref{tab:more_models} shows, within the same model architecture, as the number of parameters increases and the model complexity grows (e.g., InternVL-2), the performance generally improves, including the pre-trained version. These models also exhibit more effective learning, especially when provided with synthesized data generated by \ourmethod, which makes the learning process more efficient. Moreover, Idefics-3 shows notable improvement over its predecessor, Idefics-2, indicating that the newer version has a better visual reasoning. When comparing mPLUG-DocOwl-1.5 with models like InternVL-2, Idefics-2, and Idefics-3, it demonstrates relatively lower performance. This could be attributed to the training objective of mPLUG-DocOwl-1.5, which focuses on multi-grained text recognition and parsing, potentially resulting in weaker performance when visual reasoning is required. Nevertheless, this model still outperforms LLaVA-1.6-7B overall, which might be due to the simpler structure of the LLaVA-1.6 family. Finally, Phi-3.5-Vision-Instruct, despite having fewer parameters compared to other models, performs competitively with other models and surpasses LLaVA-1.6-7B in performance.

\section{Human Evaluation Details}
\label{appendix:human_evaluation_details}

To facilitate a rigorous human evaluation of our answer validation component, we created a Google Form to recruit participants willing to contribute to our evaluation. We shared this form widely and will acknowledge the contributions of participating individuals in the acknowledgment section of the paper's camera-ready version.

After registration, participants were divided into four batches (three participants per batch, each assigned 25 samples, 100 in total) and given access to a custom evaluation app, shown in Figure \ref{fig:eval_app}, to review the samples in their assigned batch. This application was designed to streamline the evaluation process and ensure consistency across participants. For each question, participants could review the question text, the associated Wikipedia pages, and the generated answers from two methods—one method utilizing the answer validation component and the other without it. To minimize user bias, the application randomly alternated the positioning of the methods' answers (labeling them as “Answer A” and “Answer B”) so that users could not develop a tendency to select one model over the other based on position alone. After examining the question and relevant Wikipedia content, users were asked to select one of four options to indicate their assessment of answer accuracy: (1) Answer A is correct, (2) Answer B is correct, (3) both answers are correct, or (4) neither answer is correct. 

In addition to these selections, participants had the option to provide a brief rationale for their choices. Although they have not been investigated for this research, these optional feedbacks were encouraged, as they offer valuable insights for qualitative analysis and potential future improvements in answer validation accuracy. The combination of structured and open-ended responses enhances the robustness of our evaluation and offers a more comprehensive view of user judgments, which we may explore in future iterations of our data synthesis methodology.

The evaluators had diverse academic and professional backgrounds, including graduate students in computer science, data science researchers, and software engineers with experience in NLP and machine learning. All evaluators were proficient in English and had prior familiarity with Wikipedia-style content and fact-based question answering tasks. This diversity contributed to reliable judgment across a wide range of topics and ensured that participants had the necessary background to assess factual correctness and relevance accurately. In total, twelve individuals participated in the evaluation: seven men and five women.

\section{Additional Statistics \& Information on \benchmark}
\label{appendix:benchmark_info}

\begin{figure}[!h]
    \centering
    \includegraphics[width=\linewidth]{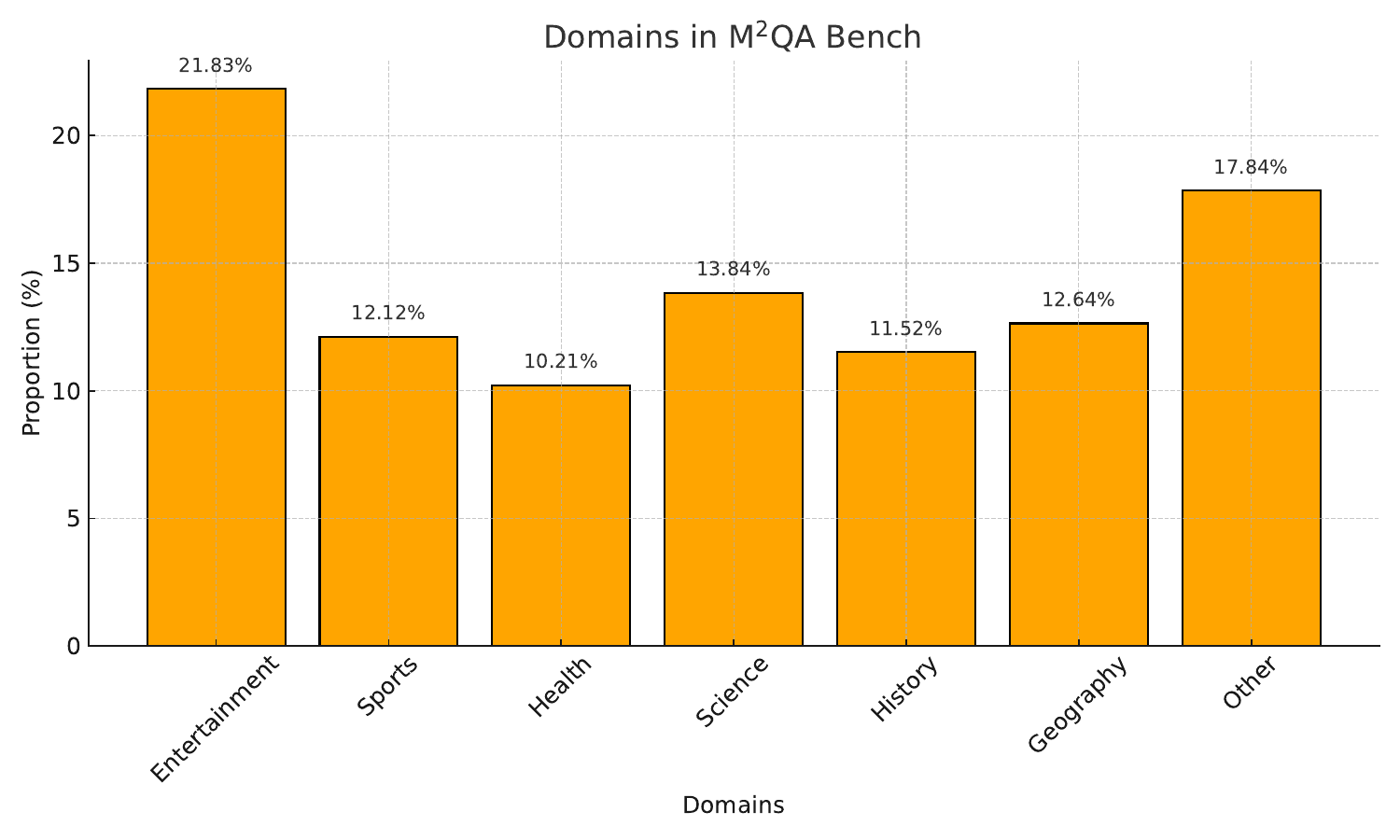}
    \caption{Distribution of domains in \benchmark.}
    \label{fig:domains}
\end{figure}

\begin{figure}[!h]
    \centering
    \includegraphics[width=\linewidth]{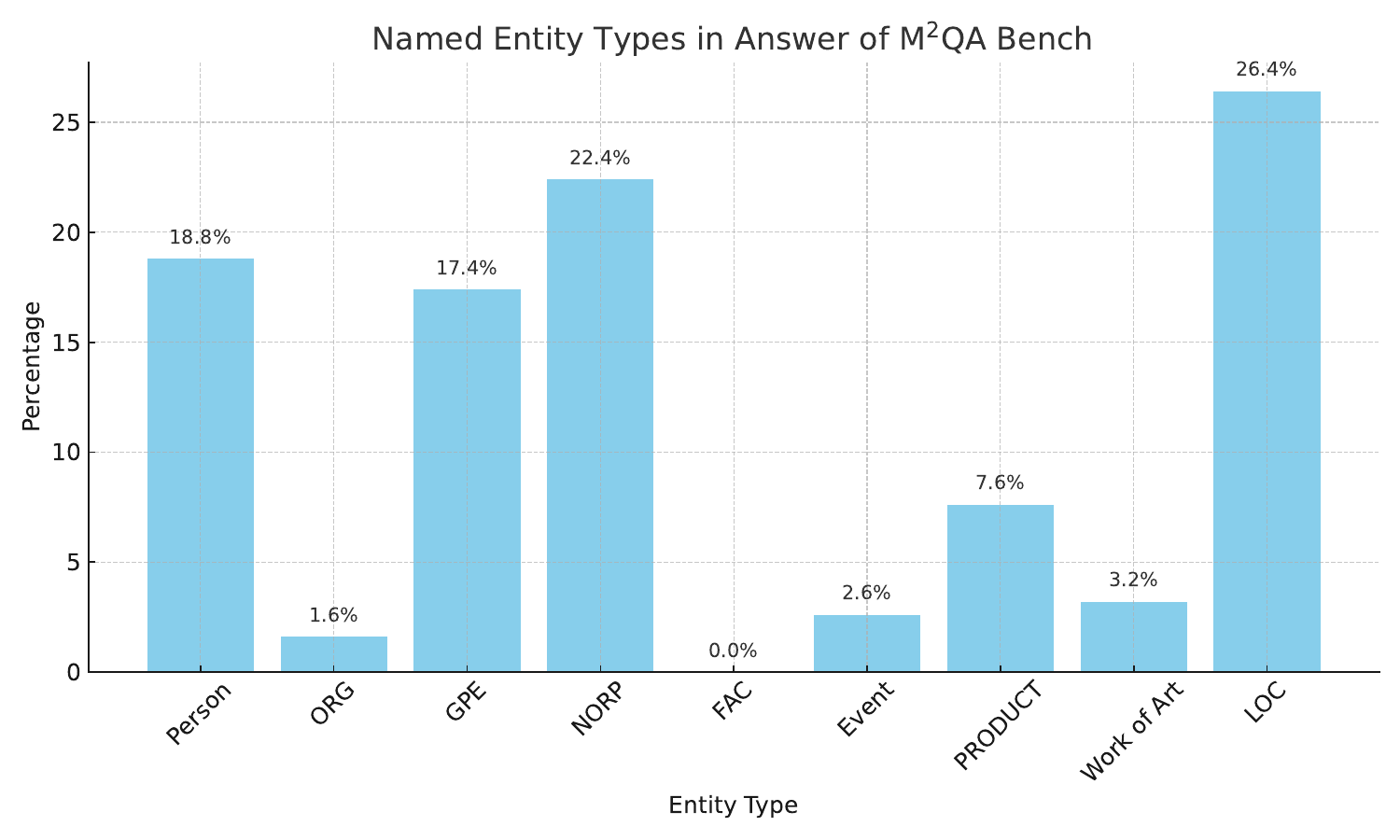}
    \caption{Distribution of named entities in answers in \benchmark.}
    \label{fig:nes}
\end{figure}

As illustrated in Figure \ref{fig:domains}, \benchmark\ encompasses a diverse range of domains. Additionally, the answers span various types of named entities, including people, products, works of art, and more. Figure \ref{fig:nes} presents the distribution of named entities found in the answers.

\begin{figure}[!h]
    \centering
    \includegraphics[width=\linewidth]{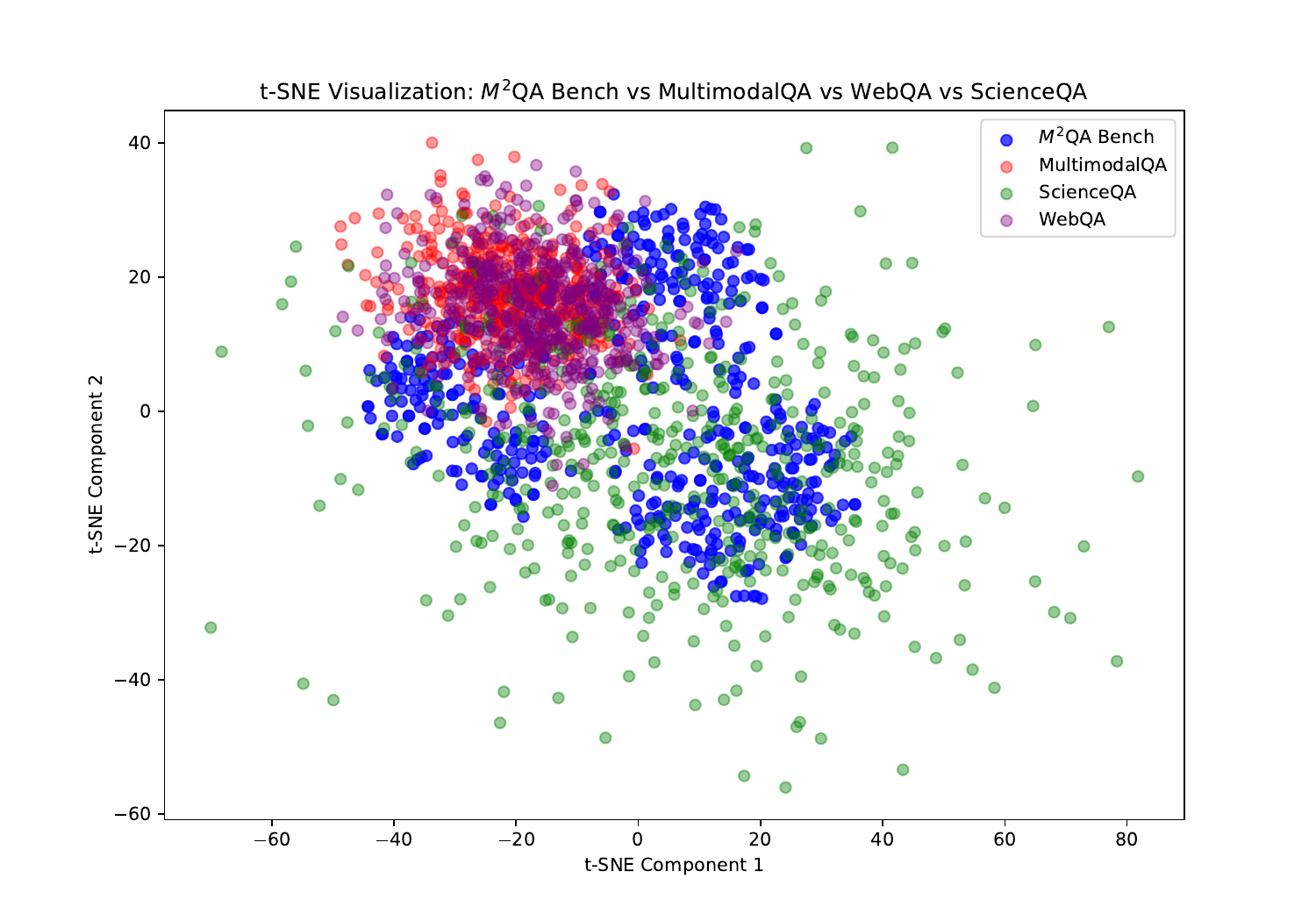}
    \caption{2D t-sne visualization of ModernBERT embeddings of questions between \benchmark, WebQA, MultimodalQA, and ScienceQA.}
    \label{fig:embeddings}
\end{figure}

To further examine the diversity of questions in our benchmark—which also reflects the overall characteristics of the data generated by \ourmethod—we conducted a 2D t-SNE analysis of question embeddings using ModernBERT \citep{warner2024smarter}. We sampled 500 questions each from \benchmark, MMQA, WebQA, and ScienceQA \citep{NEURIPS2022_11332b6b}. ScienceQA serves as a fully human-authored dataset, while MMQA and WebQA primarily use templated questions. As shown in Figure \ref{fig:embeddings}, MMQA and WebQA display the least diversity. In contrast, \benchmark, which includes questions generated from \ourmethod, demonstrates greater similarity to human-generated data, reflecting a reduced domain gap and improved diversity compared to MMQA and WebQA.

\begin{table*}[!t]
    \centering
    \resizebox{0.9\linewidth}{!}{%
    \begin{tabular}{l|c|c|c}
    \hline
    \textbf{Model} & \textbf{PMC-VQA (Acc)} & \textbf{SPIQA (F1)} & \textbf{LegalBench (Acc)} \\
    \hline
    GPT-4o (3-shot) & 57.13 & 64.84 & 77.67 \\
    InternVL2-8B (FT on 5k) & 65.72 & 77.91 & 83.45 \\
    \hline
    \end{tabular}
    }
    \caption{Cross-domain evaluation on health (PMC-VQA), science (SPIQA), and law (LegalBench). Fine-tuning InternVL-2-8B with 5k \ourmethod\ samples substantially improves performance compared to GPT-4o (3-shot) and pretrained InternVL-2-8B baselines.}
    \label{tab:other_domain_benchmarks}
\end{table*}

\section{\ourmethod\ Generalizability to Other Domains}
\label{appendix:generalizability}

\ourmethod\ can generate domain-specific synthetic data using just three in-context examples, enabling even small LVLMs to handle specialized multimodal multihop QA. As shown in Figure~\ref{fig:domains}, \ourmethod's data (including \benchmark) spans a wide range of domains. To further assess its generalizability, we trained InternVL-2-8B on 5k synthesized samples and evaluated it across three out-of-domain benchmarks: the health-related \textbf{PMC-VQA}~\citep{zhang2023pmc}, the scientific benchmark \textbf{SPIQA} \cite{pramanick2024spiqa}, and the legal reasoning dataset \textbf{LegalBench} \cite{guha2023legalbench}. 

Table~\ref{tab:other_domain_benchmarks} compares performance against GPT-4o (3-shot prompting) and the InternVL-2-8B fine-tuned on 5k samples generated by \ourmethod\ for each specific domain. The fine-tuned model consistently outperforms GPT-4o across health, science, and law, showing clear improvements in every case. This demonstrates that \ourmethod\ enables models to generalize effectively beyond the training distribution and can be readily adapted to build strong domain-expert systems across diverse fields.

\section{\benchmark\ Examples}
\label{appendix:samples}

\begin{figure*}[!ht]
    \centering
    \includegraphics[width=\textwidth]{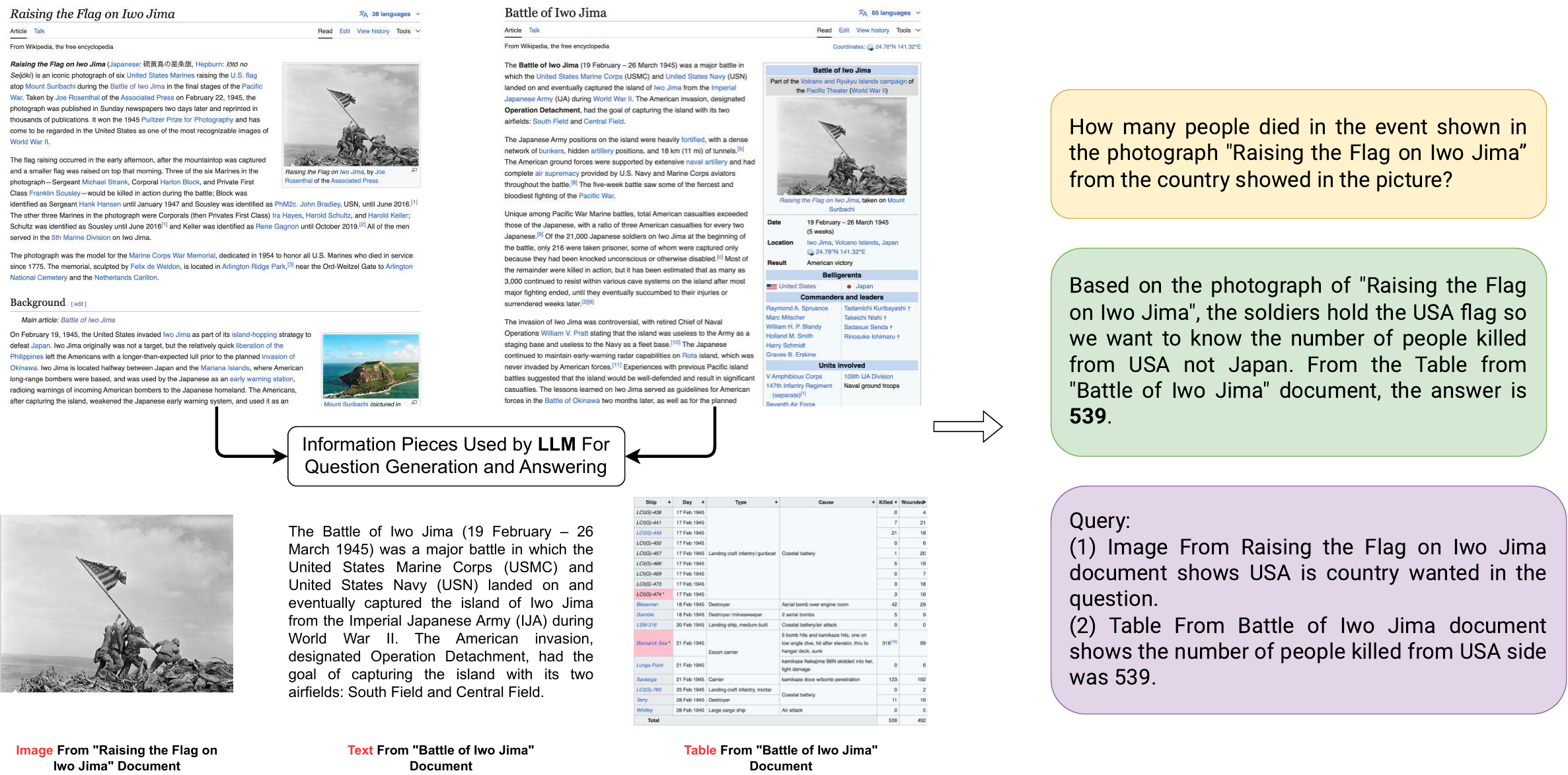}
    \caption{Multimodal and multihop reasoning example from \benchmark\ where the model answers a question about the photograph "Raising the Flag on Iwo Jima" by synthesizing information from linked documents through a hyperlink, leveraging both visual and tabular data to determine the number of casualties from the USA.}
    \label{fig:sample_1}
\end{figure*}

\begin{figure*}[!ht]
    \centering
    \includegraphics[width=\textwidth]{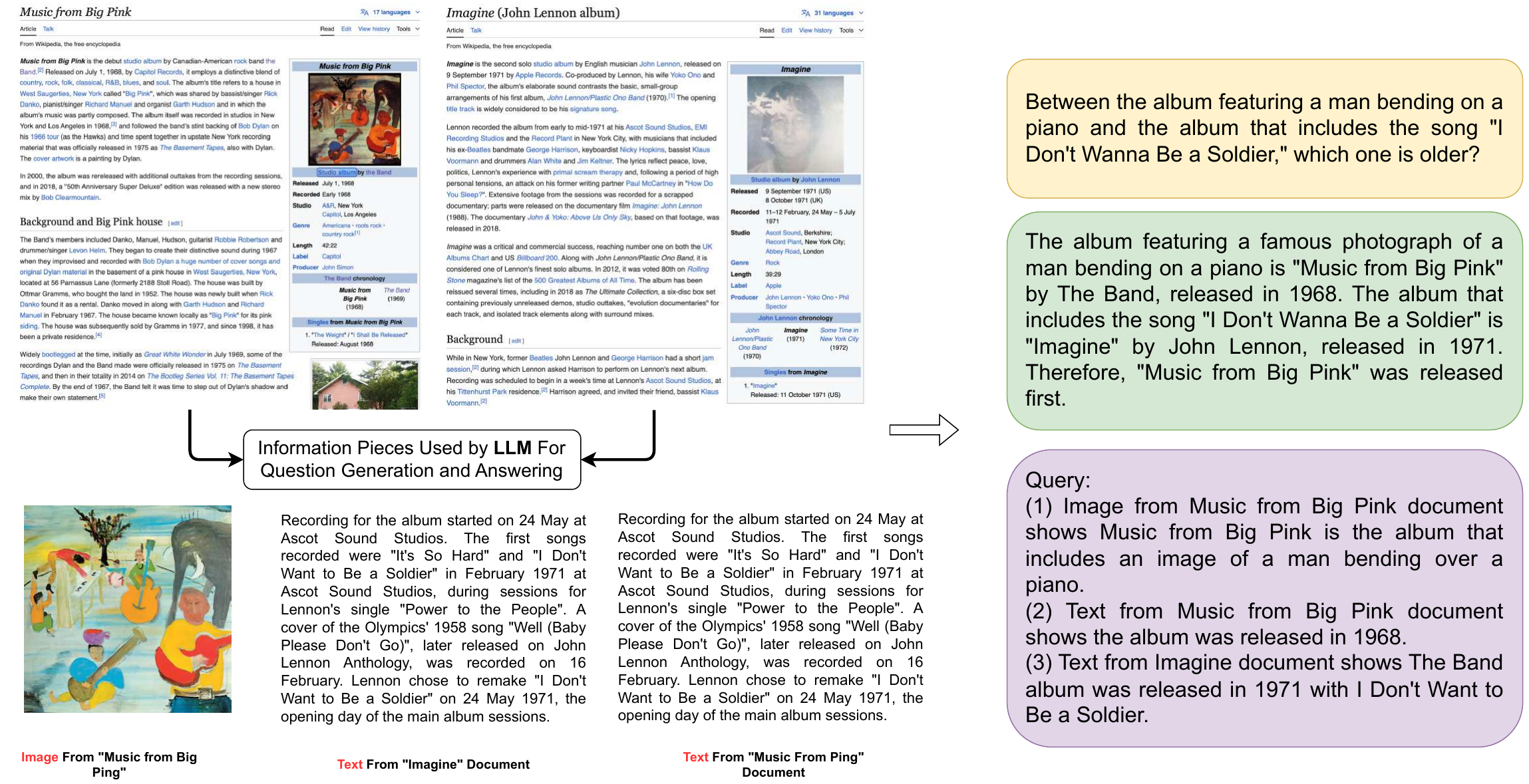}
    \caption{Multimodal multihop reasoning example from \benchmark\ where the model compares the release dates of two albums, "Music from Big Pink" and "Imagine," using textual and visual cues. The documents are connected through their shared topic, "music," and the answer is determined as the title of the earlier-released album.}
    \label{fig:sample_2}
\end{figure*}

\ourmethod\ uses LVLMs to generate multimodal and multihop questions based on the given documents and evaluate their answers. These samples aim to emulate few-shot examples typically provided to guide the model's behavior in a structured and relevant manner.

In some cases, the questions focus on understanding facts from different modalities—such as images, text, and tables—within the grouped documents and finding the answer from one of them. For example, in the case of the question shown in Figure \ref{fig:sample_1}:

\begin{quote}
\textit{How many people died in the event shown in the photograph ``Raising the Flag on Iwo Jima'' from the country shown in the picture?}
\end{quote}
LVLM is tasked with combining information from two documents: \textit{Raising the Flag on Iwo Jima} and \textit{Battle of Iwo Jima}. Here, the hyperlink between the two documents served as the connection between two docments. The model identifies that the photograph depicts American soldiers (based on the USA flag) and cross-references the table from the \textit{Battle of Iwo Jima} document to determine that 539 people from the USA were killed.  This demonstrates how the model synthesizes information across modalities to form an accurate response. Afterward, the model generates queries, serving as a step-by-step guide to extract relevant information from the documents. Using the extracted snippets, it then answers the question. For instance, the model would need to locate the image \textit{Raising the Flag on Iwo Jima} to determine the country mentioned in the question, which is the USA. Next, by referencing the table in the \textit{Battle of Iwo Jima} document, it provides the final answer.

In other cases, the questions involve comparing elements between objects in two different documents, where the answer is typically the title of one of the documents provided. For example, the question shown in Figure \ref{fig:sample_2}:

\begin{quote}
\textit{Which album was released first: the one featuring a famous photograph of a man bending on a piano or the album that includes the song ``I Don't Wanna Be a Soldier''?}
\end{quote}
requires the model to compare temporal information across two documents: \textit{Music from Big Pink} and \textit{Imagine}. The model identifies that \textit{Music from Big Pink}, featuring a photograph of a man bending on a piano, was released in 1968, while \textit{Imagine}, containing the song ``I Don't Wanna Be a Soldier,'' was released in 1971. Therefore, the answer is \textit{Music from Big Pink}. In this case, the documents were connected through their shared topic, music. The query generation in this example is similar to the first but differs slightly, as three information snippets are key to answering the question, making the query three steps long.

\section{Synthesizing Data vs. Paraphrasing Existing Human Annotated Datasets}
\label{appendix:paraphrasing}

\begin{figure}[!t]
    \centering
    \includegraphics[width=\linewidth]{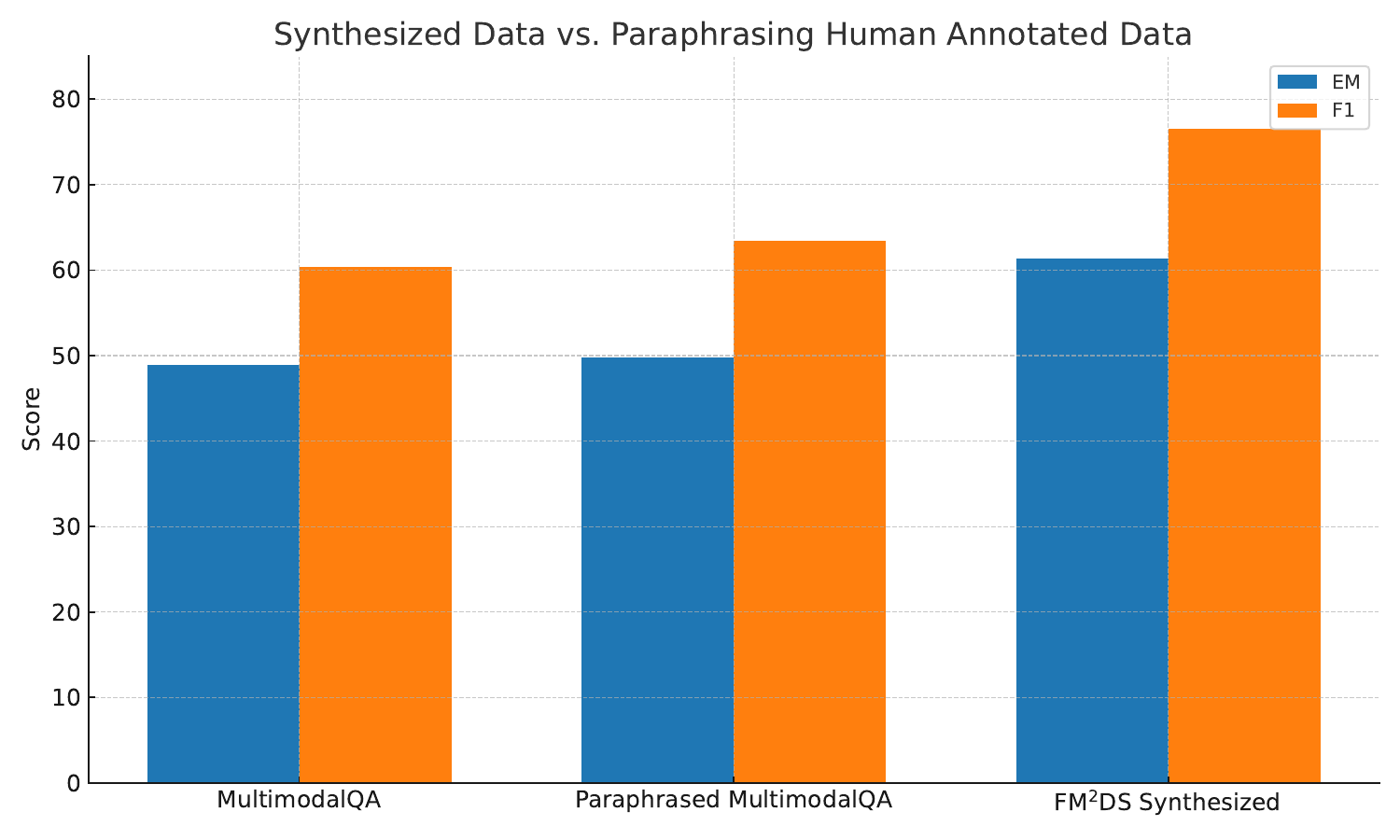}
    \caption{Performance comparison of InternVL-2-8B trained on 1k samples from three settings: original MultimodalQA, paraphrased MultimodalQA (reworded using GPT-4o), and fully synthesized data from \ourmethod. While paraphrasing existing questions yields only modest gains, our synthesized samples lead to significantly higher performance, highlighting the value of generating diverse and structurally novel multihop multimodal questions.}
    \label{fig:paraphrasing-results}
\end{figure}

Paraphrasing questions from existing datasets introduces surface-level linguistic changes but preserves the original semantic intent and reasoning pathways, offering only marginal improvements in model training. In contrast, the data synthesized by \ourmethod\ is intentionally crafted to introduce diverse question structures, span multiple domains, and require varied types of reasoning, pushing models toward more comprehensive multimodal understanding. To compare these approaches, we trained InternVL-2-8B using 1k samples from three settings: (I) the original MultimodalQA dataset, (II) a paraphrased version of MultimodalQA where questions were reworded using GPT-4o with the prompt "Please paraphrase the following question: [Question]" and (III) synthesized samples generated by \ourmethod. For all conditions, we used full Wikipedia documents as sources. Figure~\ref{fig:paraphrasing-results} presents the results, showing that while paraphrasing provides a slight improvement, synthesizing new, high-quality samples with \ourmethod\ leads to a substantial performance gain.

\begin{table*}[!h]
    \centering
    \begin{tabular}{l|c|c}
    \hline
    \textbf{Stage} & \textbf{Average Rejection Rate} & \textbf{Average Rejection Rate} \\
    \hline
    Question Validation & 131 & 11.58\% \\
    Answer Validation & 76 & 7.06\% \\
    Query Validation & 58 & 5.48\% \\
    \hline
    \end{tabular}
    \caption{Key statistics of the proposed multimodal multihop question answering benchmark.}
    \label{tab:stage_stats}
\end{table*}

\section{Statistics on Usages of Each Validation Stage of \ourmethod}
\label{appendix:each_stage_stat}

As described in Section \ref{sec:method} and illustrated in Figure \ref{fig:pipeline}, \ourmethod\ incorporates multiple validation stages to enhance data quality. It is essential to analyze how frequently each stage rejects the initially generated outputs. Table \ref{tab:stage_stats} presents statistics based on generating 1,000 examples using GPT-4o. Among the stages, question validation has the highest rejection rate, suggesting that this step is the most challenging. This may be because generating a question requires the model to synthesize all relevant knowledge and fully grasp the context. In contrast, answer validation benefits from the guidance provided by the question, making the task relatively easier. Query validation appears to be even more straightforward, as it primarily involves formatting the reasoning steps, something the model has effectively done during answer generation. Additionally, the use of question-specific image captions during answer generation likely contributes to a lower error rate by helping the model locate the correct information only text modality.

\section{Qualitative Analysis}
\label{appendix:FT_performance}

In the qualitative analysis, we compared three critical factors influencing model responses: model architecture, fine-tuning (FT) dataset (real samples or synthesized samples), and model size. To examine the effects of model architecture and FT dataset, we used InternVL-2-8B, LLaVA-1.6-7B, and Idefics-2-8B, fine-tuning them on both real and synthetic data generated by \ourmethod. For analyzing the impact of model size, all versions of InternVL-2 were trained on the synthetic data. All of the mentioned models were fine-tuned on 5k samples.

This analysis was conducted for 100 samples from each of the following benchmarks: (1) \benchmark, (2) MultiModalQA, and (3) WebQA. The results are presented in Tables \ref{tab:model_performance_stat_our}, \ref{tab:model_performance_stat_multimodalqa}, and \ref{tab:model_performance_stat_webqa}. The responses generated by different models were analyzed across these datasets, focusing on the following metrics:
\begin{enumerate}
    \item Model accuracy using the exact match (EM) metric.
    \item Hallucination rate, corresponding to instances where the model generated wrong answer based on its pre-trained knowledge instead of the provided document.
    \item Model accuracy with EM metric for samples including image modality (may include other modalities).
    \item Model accuracy with EM metric for samples including table modality (may include other modalities).
    \item Model accuracy with EM metric for samples including both image and table modalities.
\end{enumerate}

For WebQA, which only incorporates text and image modalities, the last three metrics were not applicable. Additionally, the distribution of modalities across samples for MultiModalQA and \benchmark\ was as follows:
\begin{itemize}
    \item \textbf{\benchmark{}:} 66 samples included image modality, 62 samples included table modality, and 28 samples included both image and table modalities.
    \item \textbf{MultiModalQA:} 61 samples included image modality, 54 samples included table modality, and 15 samples included both image and table modalities.
\end{itemize}

\begin{table*}[!t]
    \centering
    \resizebox{\textwidth}{!}{%
    \begin{tabular}{l|c|ccccc}
    \textbf{Model}         & \textbf{Trained On} & \textbf{EM $\uparrow$} & \textbf{Hallucination $\downarrow$} & \textbf{EM (Table)} & \textbf{EM (Image)} & \textbf{EM (Image\&Table)} \\ \hline
    \textbf{InternVL-2-8B}  & Real                & 0.43                & 0.67                     & 0.56                   & 0.51                   & 0.46                          \\
    \textbf{InternVL-2-8B}  & Synth               & \textbf{0.57}                & \textbf{0.51 }                    & \textbf{0.77 }                  & \textbf{0.68 }                  & \textbf{0.64 }                         \\ \hline
    \textbf{Idefics-2-8B}   & Real                & 0.39                & 0.72                   & 0.54                   & 0.44                   & 0.43                          \\
    \textbf{Idefics-2-8B}   & Synth               & \textbf{0.55}                & \textbf{0.68}                     & \textbf{0.71}                   & \textbf{0.62}                   & \textbf{0.53}                          \\ \hline
    \textbf{LLaVA-1.6-7B}  & Real                & 0.35                & 0.71                     & 0.43                   & 0.39                   & 0.32                         \\
    \textbf{LLaVA-1.6-7B}  & Synth               & \textbf{0.47}                & \textbf{0.51}                     & \textbf{0.61}                   & \textbf{0.46}                   & \textbf{0.39}                           \\ \hline
    \textbf{InternVL-2-26B} & Synth               & 0.61                & 0.54                     & 0.87                   & 0.74                   & 0.75                          \\
    \textbf{InternVL-2-40B} & Synth               & 0.64                & 0.5                     & 0.89                   & 0.8                   & 0.78                          \\
    \textbf{InternVL-2-76B} & Synth               & \textbf{0.72}                & \textbf{0.28}                      & \textbf{0.98}                   & \textbf{0.95}                   & \textbf{0.92}                         
    \end{tabular}
    }
    \caption{Performance of models fine-tuned on real vs. synthesized data on \benchmark. EM scores and hallucination rates are computed on filtered data (hallucination = hallucinated responses / incorrect answers). $\uparrow$ indicates higher is better (EM), and $\downarrow$ indicates lower is better (hallucination). EM (Table) and EM (Image) may overlap with other modalities. Larger models and those trained on synthesized data achieve higher EM and lower hallucination, with table questions generally easier than image ones.}
    \label{tab:model_performance_stat_our}
\end{table*}

\begin{table*}[!t]
    \centering
    \resizebox{\textwidth}{!}{%
    \begin{tabular}{l|c|ccccc}
    \textbf{Model}               & \textbf{Trained On} & \multicolumn{1}{c}{\textbf{EM $\uparrow$}} & \multicolumn{1}{c}{\textbf{Hallucination $\downarrow$}} & \multicolumn{1}{c}{\textbf{EM (Table)$\uparrow$}} & \multicolumn{1}{c}{\textbf{EM (Image)$\uparrow$}} & \multicolumn{1}{c}{\textbf{EM (Image \& Table)$\uparrow$}} \\ \hline
    \textbf{InternVL-2-8B}   & Real                & 0.66                            & 0.62                                       & 0.7                                     & 0.72                                    & 0.53                                             \\
    \textbf{InternVL-2-8B}  & Synth               & \textbf{0.68}                            & \textbf{0.41}                                       & \textbf{0.75}                                    & \textbf{0.78}                                    & \textbf{0.67}                                             \\ \hline
    \textbf{Idefics-2-8B}    & Real                & 0.61                            & 0.64                                       & 0.61                                    & 0.63                                    & 0.33                                             \\
    \textbf{Idefics-2-8B}   & Synth               & \textbf{0.64}                            & \textbf{0.53}                                       & \textbf{0.67}                                    & \textbf{0.69}                                    & \textbf{0.47}                                             \\ \hline
    \textbf{LLaVA-1.6-7B}   & Real                & 0.59                            & 0.66                                       & 0.57                                    & 0.56                                    & 0.2                                              \\
    \textbf{LLaVA-1.6-7B}  & Synth               & \textbf{0.62}                            & \textbf{0.39}                                       & \textbf{0.64}                                    & \textbf{0.61}                                    & \textbf{0.33}                                             \\ \hline
    \textbf{InternVL-2-26B} & Synth               & 0.7                             & 0.37                                       & 0.79                                    & 0.85                                    & 0.8                                              \\
    \textbf{InternVL-2-40B} & Synth               & 0.74                            & 0.35                                       & 0.85                                    & 0.89                                    & 0.87                                             \\
    \textbf{InternVL-2-76B} & Synth               & \textbf{0.8}                             & \textbf{0.15}                                       & \textbf{0.93}                                   & \textbf{0.94}                                    & \textbf{0.93}                                            
    \end{tabular}
    }
    \caption{Performance of models fine-tuned on real vs. synthesized data on MultimodalQA. EM scores and hallucination rates are computed on filtered data (EM(Table) = EM on samples with table modality). $\uparrow$ means higher is better (EM), $\downarrow$ means lower is better (hallucination). EM (Table) and EM (Image) may overlap with other modalities. Larger models and those trained on synthesized data achieve higher EM with fewer hallucinations, with image-based questions generally easier than table-based ones.}
    \label{tab:model_performance_stat_multimodalqa}
\end{table*}

\begin{table*}[!h]
    \centering
    \begin{tabular}{l|c|cc}
    \textbf{Model}         & \textbf{Train On} & \textbf{EM $\uparrow$} & \textbf{Hallucination $\downarrow$} \\ \hline
    \textbf{InternVL-2-8B}  & Real              & 0.76          & 0.41                     \\
    \textbf{InternVL-2-8B}  & Synth             & \textbf{0.78}          & \textbf{0.32}                      \\ \hline
    \textbf{Idefics-2-8B}   & Real              & 0.72          & 0.48                     \\
    \textbf{Idefics-2-8B}   & Synth             & \textbf{0.75}          & \textbf{0.36}                     \\ \hline
    \textbf{LLaVA-1.6-7B}  & Real              & 0.70          & 0.53                     \\
    \textbf{LLaVA-1.6-7B}  & Synth             & \textbf{0.74}          & \textbf{0.34}                      \\ \hline
    \textbf{InternVL-2-26B} & Synth             & 0.81          & 0.21                      \\
    \textbf{InternVL-2-40B} & Synth             & 0.82          & 0.11                      \\
    \textbf{InternVL-2-76B} & Synth             & \textbf{0.85}          & \textbf{0}                     
    \end{tabular}
    \caption{Performance of models fine-tuned on real vs. synthesized data on WebQA. Hallucination is measured as hallucinated responses over incorrect answers. $\uparrow$ means higher is better (EM), $\downarrow$ means lower is better (hallucination). Fine-tuning on synthesized data improves EM and reduces hallucination across all models, with larger models performing best. Unlike \benchmark\ and MultimodalQA, WebQA has only image and text, so EM(Image) and EM(Table) are not reported.}
    \label{tab:model_performance_stat_webqa}
\end{table*}

\begin{figure*}[!ht]
    \centering
    \includegraphics[width=0.95\textwidth]{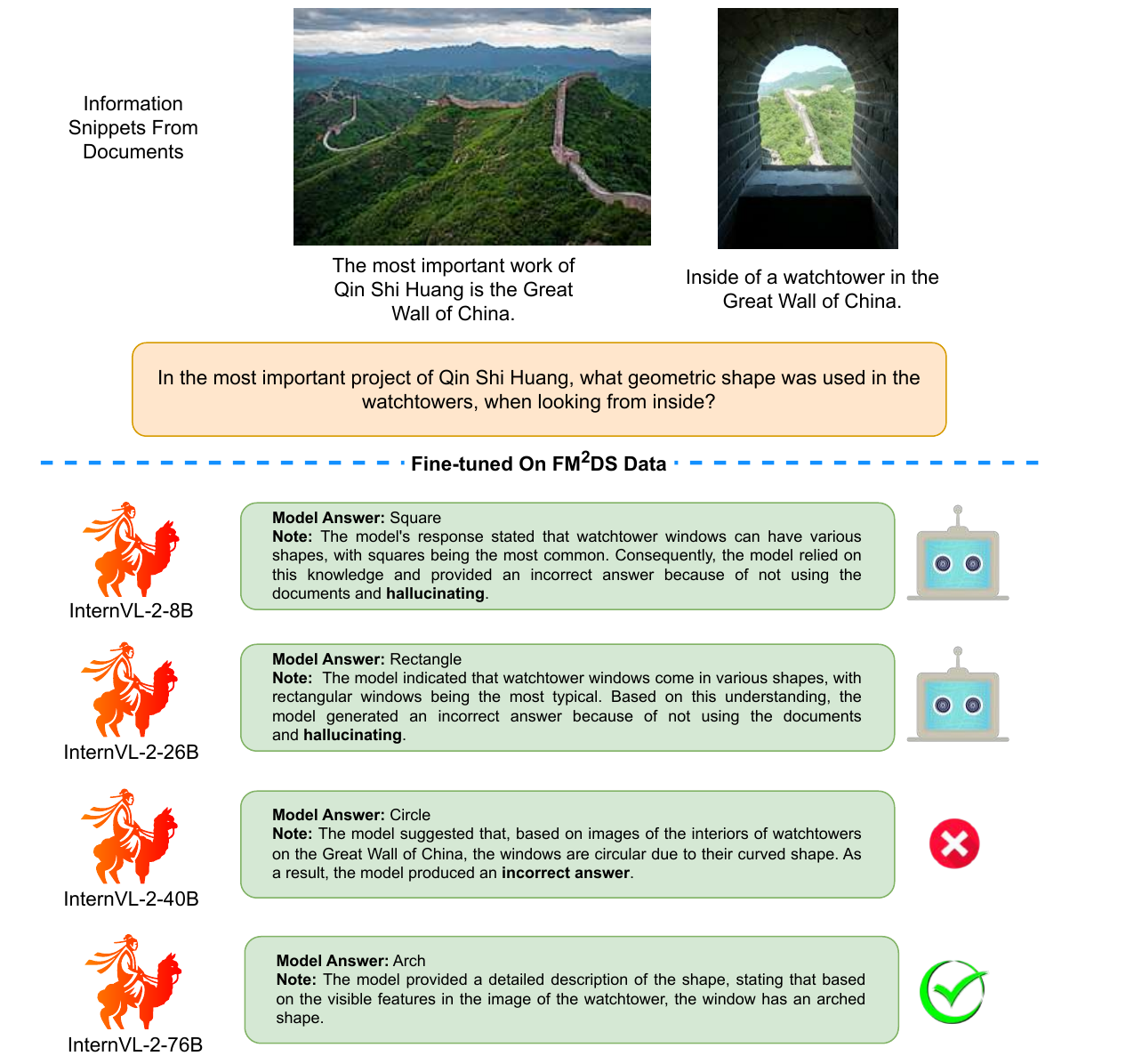}
    \caption{Responses from InternVL-2 models of various sizes (8B, 26B, 40B, and 76B) to the question: \textit{"In the most important project of Qin Shi Huang, what geometric shape was used in the watchtowers when viewed from inside?"} from \benchmark\ illustrate that in examples like this, which requires detailed visual understanding, smaller models often hallucinate, providing inconsistent answers (e.g., square, rectangle) without grounding in the provided document. Larger models, however, perform better on this task and have less hallucination.}
    \label{fig:q2}
\end{figure*}

Overall, in all benchmarks, model hallucination rates decreased as model complexity and parameter count increased, resulting in more accurate answers across all modalities (e.g., see Figure \ref{fig:q2} for an example output of these models). Larger models consistently outperformed smaller models on both modalities. Regarding synthetic data, fine-tuning on data generated by \ourmethod\ significantly reduced hallucination and improved performance across all modalities. While the hallucination rates among different model families are relatively similar, all models occasionally generate answers based on their pre-trained knowledge rather than the provided document. Fine-tuning on data generated by \ourmethod\ effectively alleviates this issue. Among the models, as shown in Table \ref{tab:main_results_formatted}, LLaVA-1.6 exhibited the poorest performance and the highest likelihood of hallucination, followed by Idefics-2, with InternVL-2 demonstrating the best performance.

\begin{figure*}[!ht]
    \centering
    \includegraphics[width=0.88\textwidth, page=1]{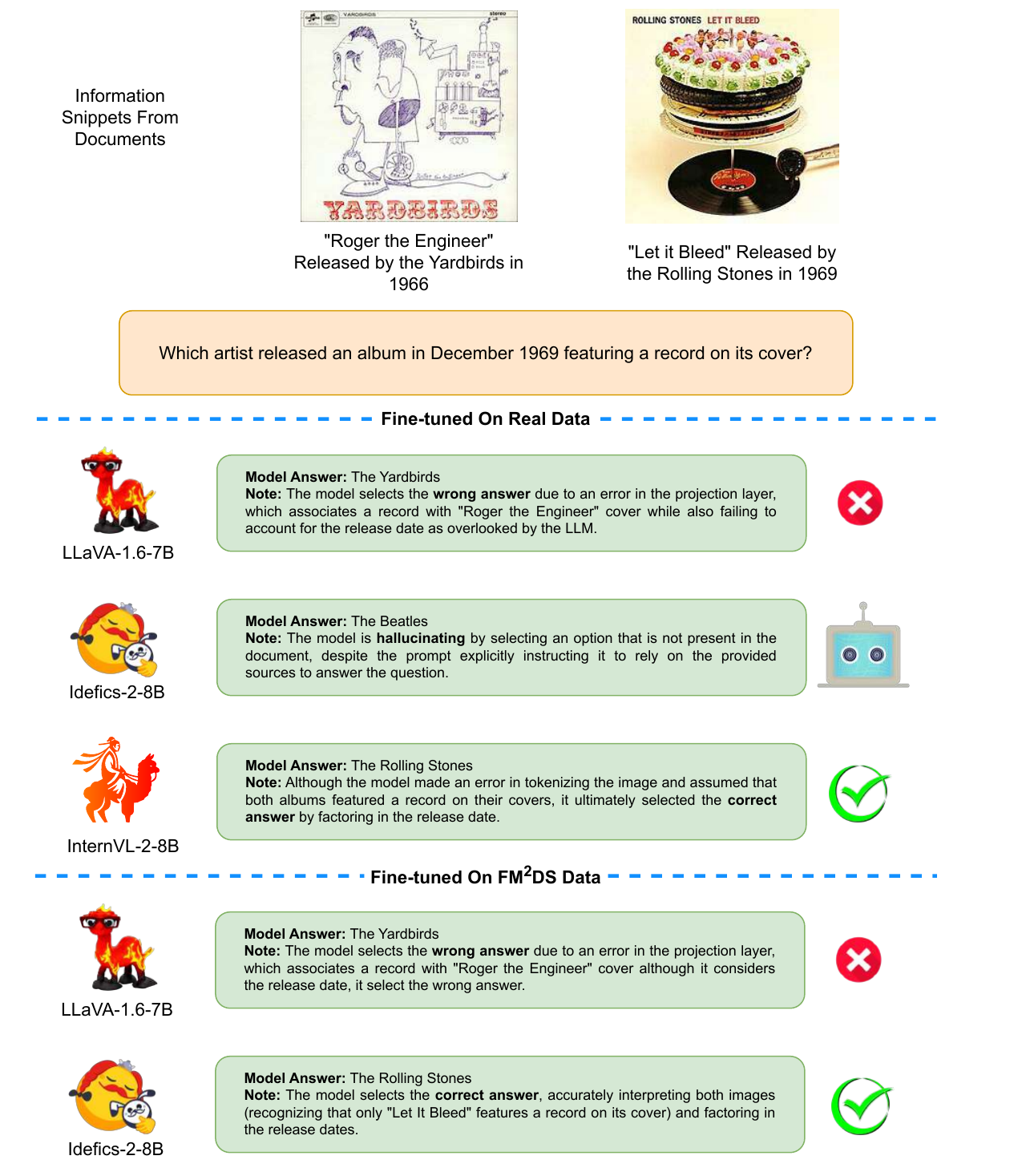}
    \includegraphics[width=0.88\textwidth, page=2]{images/Q1-cropped_compressed.pdf}
    \caption{Analysis of model responses to the question: \textit{"Which artist released an album in December 1969 featuring a record on its cover?"} from MultimodalQA dataset reveals that fine-tuning on \ourmethod\ eliminates hallucination (marked by the confused robot sign) seen in model fine-tuned on real data. This example highlights how fine-tuning improves reasoning by aligning the model's answers with both visual and textual evidence.}
    \label{fig:q1}
\end{figure*}

\begin{figure*}[!ht]
    \centering
    \includegraphics[width=0.85\textwidth, page=1]{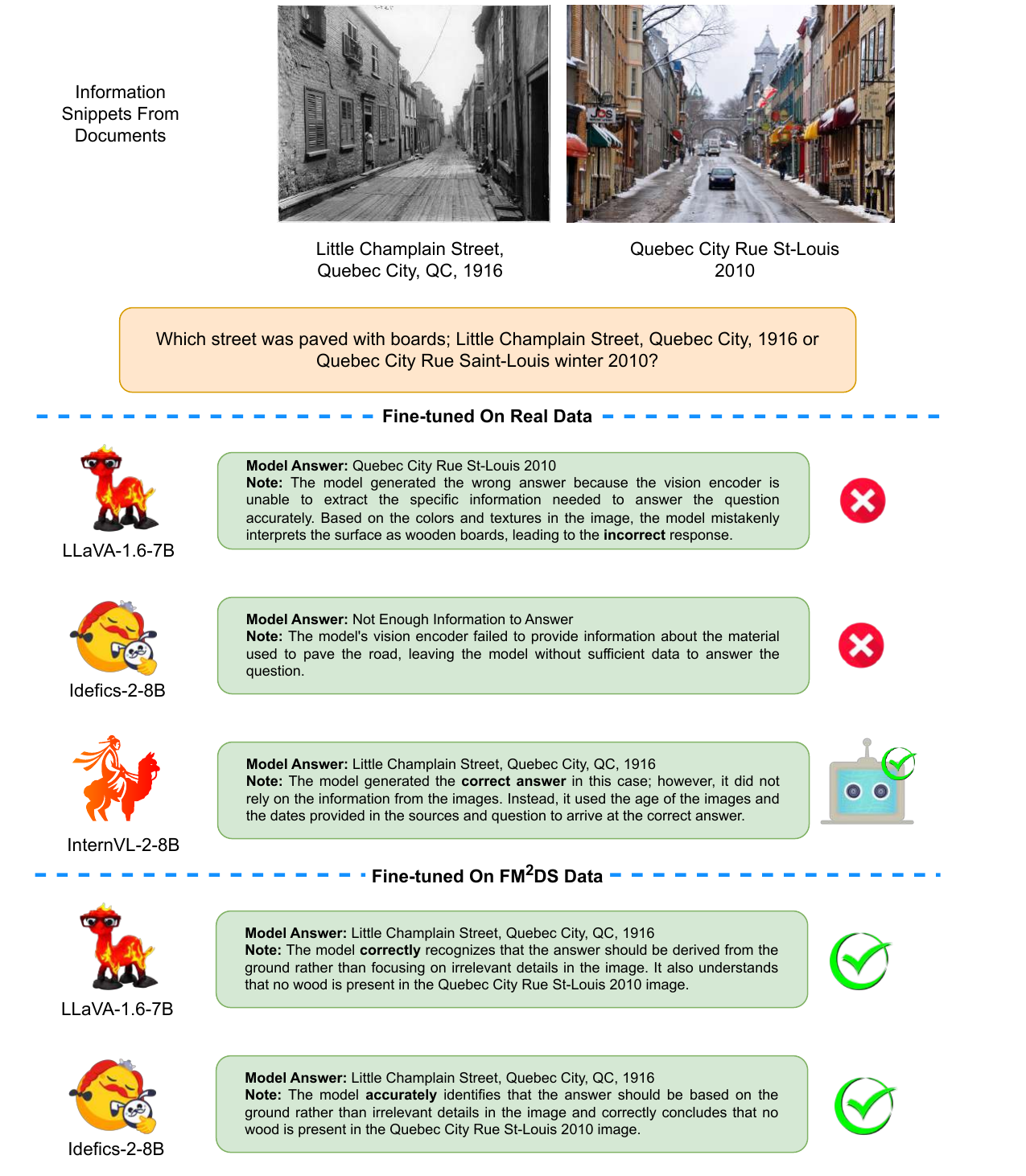}
    \includegraphics[width=0.85\textwidth, page=2]{images/Q3_compressed.pdf}
    \caption{Analysis of model responses to the question: \textit{"Which street was paved with boards; Little Champlain Street, Quebec City, 1916 or Quebec City Rue Saint-Louis winter 2010?"} from the WebQA dataset demonstrates that fine-tuning with \ourmethod\ data effectively eliminates hallucination (indicated by the confused robot sign). This example underscores fine-tuning with \ourmethod-generated data improves the model's focus on fine-grained visual details relevant to the question. Here, InternVL-2-8B fine-tuned on real data hallucinated but reached the correct answer using its pre-trained knowledge.}
    \label{fig:q3}
\end{figure*}

Regarding the effect of modalities, results from Tables \ref{tab:model_performance_stat_our} and \ref{tab:model_performance_stat_multimodalqa} suggest that the modalities themselves are not the most critical factor. 
Instead, the complexity of how the question integrates the modalities plays a more significant role. For \benchmark, models performed better when visual understanding was not required, with tables and text being the primary contributors to the results. In contrast, for MultiModalQA, models tended to perform better on image-based questions, highlighting the importance of how the question leverages the modalities. For questions involving both modalities, smaller models struggled more to produce correct answers, while larger models performed better in terms of EM. It is important to note, however, that due to the substantial difference in the number of samples containing both image and table modalities compared to those with only one modality, the reported results are not directly comparable. Refer to Figures \ref{fig:q1} and \ref{fig:q3} for the outputs of different model families fine-tuned on either real or synthesized data. Moreover, Figure \ref{fig:q2} shows outputs from different model sizes within the same family, fine-tuned on either real or synthesized data.

\end{document}